\pdfoutput=1

\documentclass[11pt]{article}

\usepackage{acl}

\usepackage{times}
\usepackage{latexsym}

\usepackage[T1]{fontenc}

\usepackage[utf8]{inputenc}

\usepackage{microtype}

\usepackage{inconsolata}

\usepackage{graphicx}
\usepackage{multirow} 
\usepackage{booktabs} 

\title{Sample Design Engineering: An Empirical Study of What Makes Good Downstream Fine-Tuning Samples for LLMs}

\author{Biyang Guo$^{1\dag}$, He Wang$^{1\dag}$, Wenyilin Xiao$^{1\dag}$\\ {\bf Hong Chen$^{2\dag}$, Zhuxin Lee$^{3}$, Songqiao Han$^{1*}$, Hailiang Huang$^{1,4*}$}
\\
        $^{1}$AI Lab, SIME, Shanghai University of Finance and Economics \\ 
        $^{2}$Ant Group, $^{3}$Guangdong Yunxi Technology \\
        $^{4}$Key Laboratory of Interdisciplinary Research of Computation and Economics, \\
        Ministry of Education, China
\\}

\begin{document}
\maketitle
\begingroup\def\thefootnote{$\dag$}\footnotetext{Equal Contribution}\endgroup
\begingroup
\def\thefootnote{$*$}
\footnotetext{\raggedright Corresponding authors, emails:}\endgroup
\begingroup\def\thefootnote{}\footnotetext{han.songqiao@shufe.edu.cn,  hlhuang@shufe.edu.cn}\endgroup
\begin{abstract}
In the burgeoning field of Large Language Models (LLMs) like ChatGPT and LLaMA, Prompt Engineering (PE) is renowned for boosting zero-shot or in-context learning (ICL) through prompt modifications. Yet, the realm of the sample design for downstream fine-tuning, crucial for task-specific LLM adaptation, is largely unexplored. This paper introduces \textbf{Sample Design Engineering} (SDE), a methodical approach to enhancing LLMs' post-tuning performance by refining input, output, and reasoning designs.
We conduct a series of in-domain (ID) and out-of-domain (OOD) experiments to assess the impact of various design options on LLMs' downstream performance, revealing several intriguing patterns that hold consistently across different LLMs. Based on these insights, we propose an integrated SDE strategy, combining the most effective options, and validate its consistent superiority over heuristic sample designs in complex downstream tasks like multi-aspect sentiment analysis, event extraction, and nested entity recognition.
Additionally, analyses of LLMs' inherent prompt/output perplexity, zero-shot, and ICL abilities illustrate that good PE strategies may not always translate to good SDE strategies. Code available at \url{https://github.com/beyondguo/LLM-Tuning}.

\end{abstract}

\section{Introduction}

The emergence of Large Language Models (LLMs) such as GPT-3 \cite{GPT-3}, PaLM \cite{chowdhery2023palm}, LLaMA \cite{touvron2023llama-1} and GPT-4 \cite{achiam2023gpt4} revolutionized natural language processing (NLP), enabling complex tasks to be tackled with a single model. This shift has profoundly broadened the range of tasks manageable by NLP models, while simultaneously consolidating the methodologies for various tasks under the unified framework of text generation. In this background, \textbf{\textit{Prompt Engineering}} (\textbf{PE}) has emerged as a key area in leveraging cutting-edge LLMs, leading to advances in applying LLMs to new tasks \cite{GPT-3}, enhancing logical reasoning \cite{wei2022COT}, and increasing task-specific accuracy \cite{Wang2023Prompt-health,wei2023chatie}, without updating model weights.

\begin{figure}
    \centering
    \includegraphics[width=0.85\linewidth]{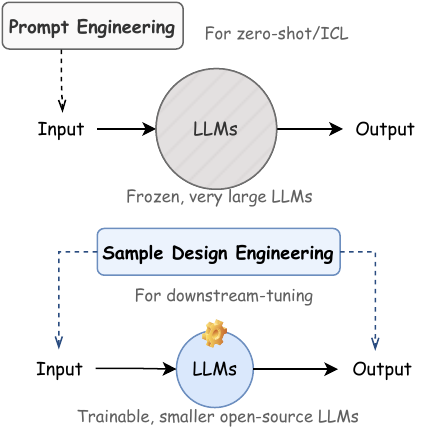}
    \caption{A simplified comparison between PE and our proposed SDE.}
    \label{fig:first-image}
\end{figure}

While numerous PE techniques have been developed for LLMs' zero-shot and in-context learning (ICL), the challenge of designing effective training samples for fine-tuning LLMs—termed \textbf{\textit{Sample Design Engineering}} (\textbf{SDE}) in this paper—remains underexplored. SDE is crucial for tailoring smaller open-source LLMs to specific requirements, especially given the complexity of training samples for downstream tasks. Figure \ref{fig:first-image} is a simplified demonstration of PE and SDE.

To address this gap, this paper undertakes a detailed and comprehensive exploration of \textbf{SDE} for LLMs' downstream fine-tuning. Our study is based on the hypothesis that the structure or elements of training samples may have a big impact on the fine-tuned LLMs. Different sample designs may make it easier or harder for the LLMs to learn, especially in scenarios where data is scarce. 


We begin by identifying a range of typical SDE options and categorizing them into three groups: \textit{input}, \textit{output }, and \textit{reasoning design options} (shown in Figure \ref{fig:Design-demo}). To reveal the impact of each SDE option, we conduct experiments on a typical downstream scenario – multi-aspect sentiment analysis (MASA), with 2 in-domain (ID) tasks and 2 out-of-domain (OOD) tasks. Different from instruction-tuning datasets like FLAN \cite{longpre2023-FLAN-data}, the MASA task involves more complicated input and output elements, making it suitable for in-depth investigation of different sample designs.
Comprehensive experiments on these 4 tasks with 6 popular open-source LLMs are undertaken to reveal how different SDE options affect downstream performances. Some interesting and thought-provoking conclusions are revealed through our experiments. For example, simply switching the position of the task instruction can make a difference; adding placeholders to unmentioned targets brings a notable performance gain, etc.

Leveraging these findings, we combine the empirically well-performing SDE options and propose an integrated SDE strategy \textbf{ES-SDE}. Extensive experiments on 3 complex downstream tasks (Nested-NER, Event Detection, and MASA) on 2 additional LLMs demonstrate that ES-SDE notably surpasses weaker SDE combination, as well as heuristic design from other studies. 
ES-SDE's robustness on different training sizes, decoding randomness or instruction variation further underscores its stable effectiveness.

In an exploratory analysis, we investigate the link between effective prompt and sample designs, via perplexity, zero-shot, and ICL analysis. Our findings suggest that a well-crafted PE strategy may not necessarily translate to a successful SDE strategy. This observation encourages further research into SDE's mechanisms, promising for enhancing LLMs' downstream applications.

\section{Background and Related Work}
\subsection{Prompt Engineering (PE)}
The effectiveness of PE methods is largely built upon the strong inherent capabilities of LLMs, with most research focusing on very large models such as GPT-3, GPT-4, PaLM, etc. (refer to \citet{sahoo2024_prompt_engineer_survey}). These models are pre-trained on extremely vast corpora, acquiring a wealth of knowledge and patterns, which enables them to directly perform complex tasks through careful prompt design. For instance, \citet{GPT-3} use carefully crafted prompts and in-context-learning (ICL) techniques to guide GPT-3 on novel tasks without training; \citet{wei2022COT} propose the Chain-of-Thought (CoT) technique that can boost the logic reasoning performance; RAG \cite{lewis2020_RAG} and CoVe \cite{dhuliawala2023_CoVe} methods are used to reduce hallucination during generation; \citet{li2023_emotion_prompt} introduce EmotionPrompt to improve LLMs' emotional intelligence.

However, these most advanced and effective LLMs are either black-box models that are only accessible via APIs, or extremely large models that are unaffordable for most companies to serve in production.
Consequently, many practitioners turn to smaller but open-source LLMs, especially 10B around models. In this situation, solely relying on PE for zero-shot or ICL inference is unable to handle many real-world complex NLP tasks. 

\subsection{Fine-tuning LLMs}
According to the different purposes, we can divide LLMs' fine-tuning into two types: \textit{instruction-tuning} (IT) and \textit{downstream-tuning} (DT)\footnote{It is also known as task tuning (TT) in some literature, like \cite{Weber2023Mind-instructions}.}. 

IT trains LLMs to comprehend and execute instructions across a range of NLP tasks, enabling predictions for new tasks \cite{wei2021finetuned-FLAN-model, mishra2022cross-insturction} with datasets like FLAN \cite{longpre2023-FLAN-data}, Self-instruct \cite{wang2023self-instruct}, Alpaca \cite{taori2023-alpaca} and HC3 \cite{guo2023-HC3}, covering tasks like such as classification, QA and translation. This is mainly applied to \texttt{base} models to enable them to follow general human instructions. DT focuses on customizing LLMs for specific, often complex, tasks in industrial applications, demanding high output stability for easier parsing and application in downstream products. An example is multi-aspect sentiment analysis, which requires detailed task instructions and outputs. Our study centers on SDE in DT scenarios, highlighting sample design challenges, but the insights may also benefit IT sample design, a topic for future exploration.

\begin{figure*}[t]
    \centering
    \includegraphics[width=\textwidth]{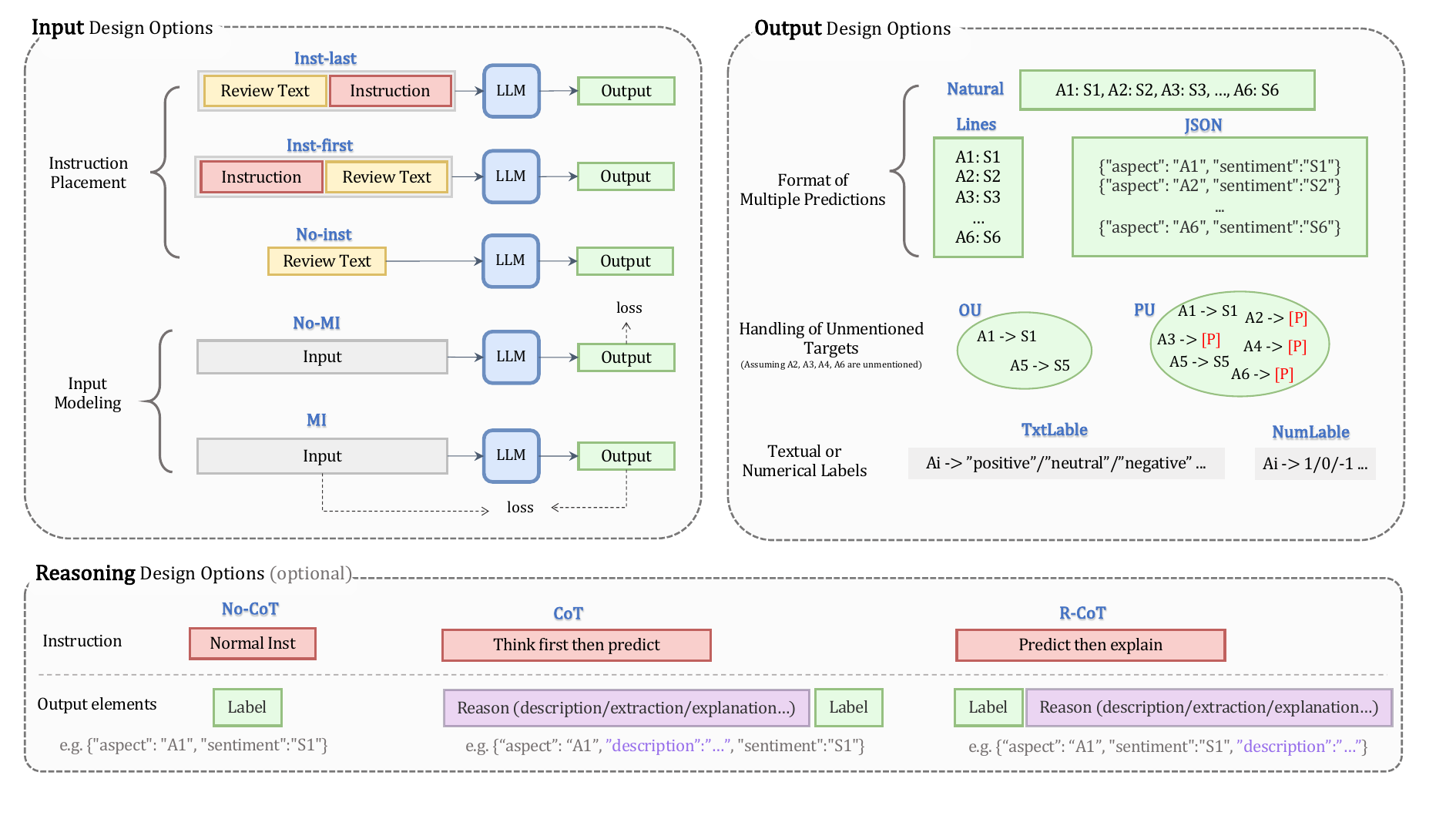}
    \caption{\small{Typical SDE options to be considered when designing downstream-tuning samples, taking the MASA task as an example. $Ai$ means aspect $i$, $Si$ means its sentiment label, \texttt{[P]} refers to placeholder tokens.}}
    \label{fig:Design-demo}
\end{figure*}

\begin{figure}[h]
    \centering
    \includegraphics[width=1\linewidth]{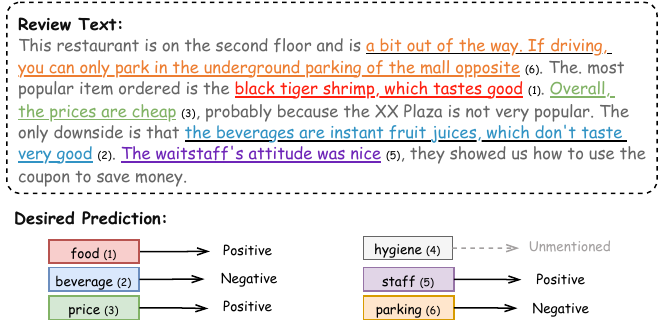}
    \caption{An example for the MASA task.}
    \label{fig:MASA-example}
\end{figure}

\subsection{Parameter-efficient fine-tuning}
The expansion of language models has made traditional full-parameter fine-tuning (FFT) less viable due to its high computational and storage demands. Parameter-efficient fine-tuning (PEFT) methods, such as prefix-tuning\cite{li2021prefix-tuning}, prompt-tuning\cite{lester2021_prompt-tuning}, p-tuning\cite{liu2023gpt_understands_P-Tuning}, and LoRA\cite{lora} provide cost-effective alternatives that retain FFT's effectiveness, gaining popularity in industrial applications. These techniques are adaptable to both IT and DT scenarios. In this research, we use the widely-used LoRA as the default fine-tuning technique. However, we believe results from our study are also applicable to other PEFT methods.

\section{Sample Design Engineering}

\subsection{Typical SDE Options}
We categorize sample design options into three aspects: \textit{input}, \textit{output}, and \textit{reasoning}. We take the Multi-Aspect Sentiment Analysis (MASA), a typical downstream task, as an example to clarify each design option for fine-tuning samples. As illustrated in Figure \ref{fig:MASA-example}, MASA requires analyzing review texts to assign sentiments to predefined aspects, while some aspects may be unmentioned. 
Figure \ref{fig:Design-demo} is an overview of different SDE options, which should be considered to design proper DT samples.

\subsubsection{Input Design Options}

$a.$ \textbf{Instruction Placement}: We explore the effect of instruction positioning relative to task text (for MASA, the review text), examining \textit{\textbf{Inst-first}} (before the task text), \textit{\textbf{Inst-last}} (after the task text). We also compare with the \textbf{\textit{No-inst}} (no instruction) option to evaluate the effectiveness of explicit instructions, as used in many previous conditional text generation tasks \cite{lewis2019bart,guo2022genius,zhang2023CTG-survey}.

\noindent$b.$ \textbf{Input Modeling}: Considering the distinction between unified sequence modeling in LLM pre-training and the explicit input/output segmentation in fine-tuning, we compare \textit{\textbf{No-MI}} that excluding input from loss calculation, akin to LLaMA2's SFT process \cite{touvron2023llama}) against \textit{\textbf{MI}} (modeling input in backpropagation).

\subsubsection{Output Design Options}
$a.$ \textbf{Multiple Predictions Formatting}: For tasks necessitating several predictions, we evaluate output formatting from less to more structured: \textit{\textbf{Natural}} (free-form text), \textit{\textbf{Lines}} (each aspect on a new line), and \textit{\textbf{JSON}} (JSON-lines for precision and explicitness).

\noindent$b.$ \textbf{Handling Unmentioned Targets}: We consider whether to omit the unmentioned (\textit{\textbf{OU}}) targets in the output, or place placeholders (\textit{\textbf{PU}}) for those targets. The placeholder tokens can be strings like "Unmentioned", "None", or "[]" according to tasks.

\noindent$c.$ \textbf{Textual or numerical labels}: By default, we use the \textbf{\textit{TxtLabel}} option for textual output labels. However, in some cases, using numbers to represent outcomes  (\textbf{\textit{NumLabel}}) may enhance prediction robustness.

\subsubsection{Reasoning Design Options}
Many tasks require reasoning, where the Chain-of-Thought (CoT) \cite{wei2022COT} has shown promise in improving LLM's reasoning in zero-shot and ICL, as well as IT scenarios \cite{kim2023cot-collection}. Yet, its impact on DT remains less studied.

We introduce the \textbf{\textit{CoT}} option for training models to "think before they predict". We use \textit{JSON} as the default output format to make the representation clearer and add a new {\color{blue}\texttt{description}} field before the {\color{blue}\texttt{sentiment}} field.
Conversely, the \textbf{\textit{R-CoT}} (Reverse-CoT) reverses these fields, enabling a "predict then explain" approach to explore CoT's mechanics further.
Note that Implementing CoT-like samples incurs additional annotation costs due to the \texttt{description} fields, making the reasoning design options task-dependent.

\subsection{Integrated SDE Strategy}
A final sample design is a combination of the above design options, which we call an \textbf{integrated SDE strategy}. This paper initially explores the impact of each individual option through extensive experimentation, leading to the proposal of an evidence-based integrated SDE strategy.

\section{Experiments \uppercase\expandafter{\romannumeral1}: Evaluating The Impact of Each SDE Option}\label{sec:Exp-1}
\subsection{Settings}

\textbf{\textit{Tasks and Datasets.}} We experiment with in-domain (ID) evaluations and out-of-domain (OOD) evaluations, for the Chinese online review MASA scenario.
The data is provided and annotated by our collaborating company, which encounters a real-world business need for the analysis of extensive customer online reviews. The data annotations come from two domains of aspects: \textbf{D1} about food, beverage, price, hygiene, staff attitude, and parking convenience and \textbf{D2} about traffic convenience, queuing, serving speed, decoration, and noise. The model needs to give a sentiment label from \{\textit{positive}, \textit{neutral}, \textit{negative}\} for each aspect, while some aspects may not occur in the review.
Based on the two domains, we construct the following 4 tasks: \\
$\bullet$ \textbf{D1$\Rightarrow$D1} and \textbf{D2$\Rightarrow$D2} are two ID evaluation tasks, where train and test sets come from the same domains; \\
$\bullet$ \textbf{D1$\Rightarrow$D2} and \textbf{D2$\Rightarrow$D1} are two OOD generalization tasks, where the model trains on one domain but tests on an unseen domain.

Considering the high cost of annotation in industries and the fact that fine-tuning LLMs requires less annotated data \cite{zhou2024lima}, we train the model with $500$ and $1,000$ samples, respectively. We use a large test set containing around $8,000$ samples to make results more stable and convincing. Dataset details see Appendix \ref{app:datasets}.

\begin{figure*}
    \centering
    \includegraphics[width=1\linewidth]{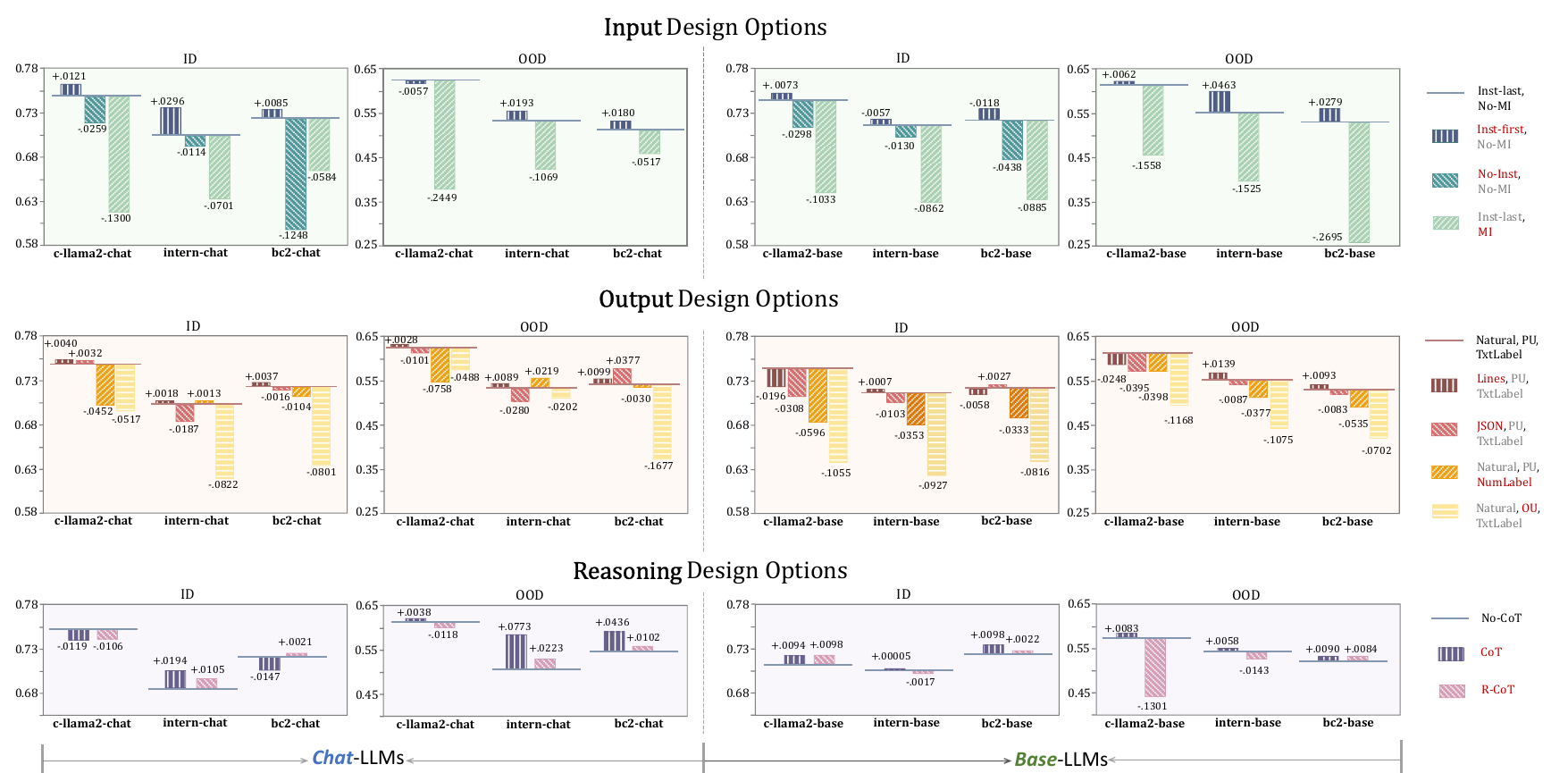}
    \caption{Sentiment analysis performances ($\kappa$) of different SDE options. Results of ID are the average of D1->D1 and D2->D2, same for OOD. The bars depict each method's relative improvement or degradation compared to the baseline, with each method differing from the baseline in only one option (colored in {\color{red}red}). Detailed results for each task see Table \ref{fig:LLaMA2-Chat_result}-\ref{fig:Baichuan2-Base_result}.}
    \label{fig:big-results-figure}
\end{figure*}

\noindent\textbf{\textit{Models.}} We utilize the following widely used open-source LLMs of 7B size of both the \textit{base} and \textit{chat} versions: 
1) \textit{chinese-llama-2-7b} (note as \textbf{c-llama2-base}) and the instruction-tuned version \textit{chinese-alpaca-2-7b} (\textbf{c-llama2-chat}) from the Chinese-LLaMA2 series \cite{cui2023efficient}, which is the vocabulary-expanded version of LLaMA2 \cite{touvron2023llama} with secondary pre-training and fine-tuning on Chinese corpus; 2) \textit{internlm-7b-base} (\textbf{intern-base}) and \textit{internlm-7b-chat} (\textbf{intern-chat}) from the InternLM series \cite{team2023internlm}, which are pretrained on trillions of high-quality tokens, performs well in Chinese and English tasks; 3) \textit{baichuan2-7b-base} (\textbf{bc2-base}) and \textit{baichuan2-7b-chat} (\textbf{bc2-chat}) from the Baichuan2 series \cite{yang2023baichuan}, one of the SOTA LLMs at the time of release. 
We use LoRA as the default efficient fine-tuning technique. Hyperparameters and other training details can be found in Appendix \ref{app:datasets}.

\noindent\textbf{\textit{Evaluation Metrics.}} We evaluate the MASA's performance from two perspectives: 1) \textbf{Sentiment analysis performance}. We use the weighted Kappa score $\kappa$ \cite{cohen1968weighted} for this measurement considering the imbalance of different aspects and the ordinal nature of sentiment labels. The weighted Kappa score allows for setting weights to enable a nuanced assessment of different classification error degrees \cite{YILMAZ2023110020}. For example, classifying "positive" as "negative" is more detrimental than classifying "positive" as "neutral," hence a higher penalty should be imposed on the former. 2) \textbf{Format adherence}, to assess the generation stability of LLMs. It's vital to have good format adherence ability for LLMs on downstream tasks so the output can be parsed successfully. We report the format-parsing error rate for this metric. Note that when calculating $\kappa$, we use relaxed parsing rules to allow some minor uncertainty of aspect/label expressions. If a certain aspect can still not be parsed correctly, this aspect is treated as "unmentioned".
The definition of $\kappa$, Kappa weight matrix, and format-parsing rules can be seen in Appendix \ref{app:metrics}.

\subsection{Experimental Results on Each Option}
We report and analyze the results from two perspectives—sentiment analysis performances, and format adherence abilities.

\subsubsection{Sentiment Analysis Performance}
We first assess the sentiment analysis performances of LLMs using different sample design options.
The comparative results of ID and OOD tasks on 3 Chat-LLMs and 3 Base-LLMs are plotted in Figure \ref{fig:big-results-figure} (full results see Table \ref{fig:LLaMA2-Chat_result} to Table \ref{fig:Baichuan2-Base_result} in Appendix \ref{app:experiment1}). Some shared and intriguing patterns are revealed from the results.

\noindent\textbf{\textit{Conclusions for Input Options:}}\\
\noindent1) \textbf{Instructions enhance DT performances}: The \textit{No-Inst} option leads to poorer performance in ID tasks and a lack of OOD generalization ability compared to \textit{Inst-first} or \textit{Inst-last} methods that incorporate instructions. This underlines the critical role of including instructions for improving both understanding and generalizability of LLMs.\\
\noindent2) \textbf{Better to place instruction first}: The \textit{Inst-first} method outperforms \textit{Inst-last} across both ID and OOD tasks for different LLMs. This demonstrates the significance of instruction placement for LLMs' tuning process. We hypothesize that this may partly be explained by the attention mechanism, see Appendix \ref{app:input}.\\
\noindent3) \textbf{Modeling input detracts from performance}: Employing the \textit{MI} approach results in worse outcomes compared to the \textit{No-MI} baselines across various models and tasks. This indicates that modeling the input part during fine-tuning may hinder the LLM's effectiveness, suggesting a cautious approach to what aspects of the task are modeled.

\begin{figure*}[h]
    \centering
    \includegraphics[width=1\linewidth]{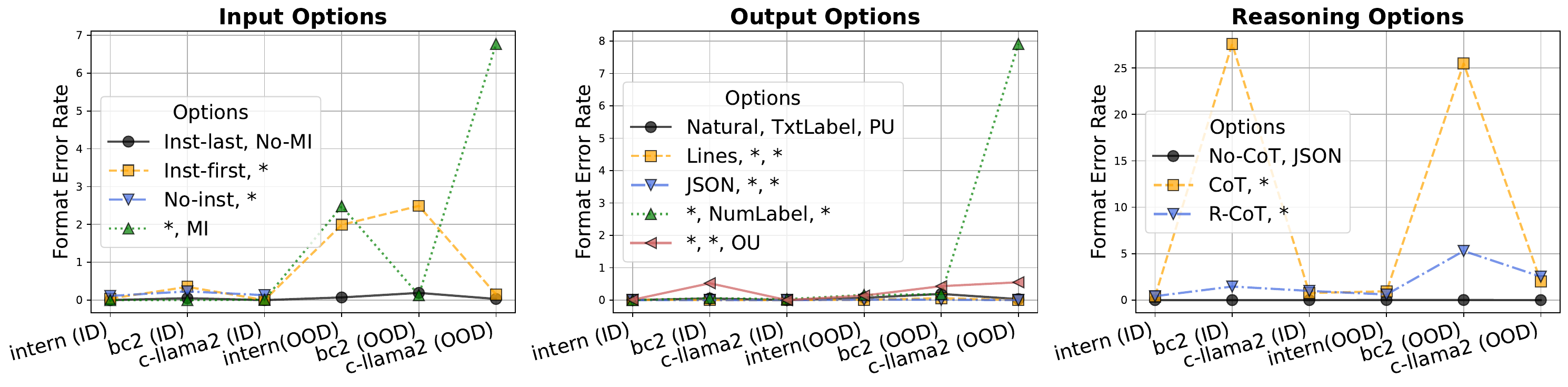}
    \caption{Format adherence performance, measured by parsing error rates (\%). '*' means same option as above.}
    \label{fig:error-rate}
\end{figure*}

\noindent\textbf{\textit{Conclusions for Output Options:}}\\
\noindent1) \textbf{\textit{Lines} is a reliable output format for multiple predictions}: The \textit{Lines} format, positioned between the \textit{Natural} and \textit{JSON} formats, demonstrates stable and high performance in sentiment analysis across various models and tasks. Its effectiveness lies in offering structured information while retaining natural language readability, making it versatile for different LLMs.\\
\noindent2) \textbf{Base-LLMs exhibit similar patterns while Chat-LLMs diverse}: Base models respond similarly to output formats, indicating consistency in their responses. In contrast, Chat models, such as bc2-chat and cllama2-chat, exhibit varied performances, suggesting differences in their SFT or RLHF data's structure. For instance, bc2-chat and cllama2-chat perform well with \textit{JSON} format, unlike intern-chat, implying a variance in the amount of structured data used in training.\\
\noindent3) \textbf{Base-LLMs favor more natural formats while Chat-LLMs can fit or bear more sophisticated formats}: Base models prefer \textit{Natural} and \textit{Lines} over \textit{JSON}. Conversely, Chat models lean towards structured formats, with \textit{Lines} and \textit{JSON}. This divergence hints at the different training backgrounds, with Chat models being more accommodating to sophisticated data formats. One more piece of evidence is that the \textit{NumLabel} option brings much more damage to the Base models than to the Chat models, which is less natural than \textit{TxtLabel}.\\
\noindent4) \textbf{Textual over numeric labels}: Switching from textual to numeric labels worsens performance, likely because numeric labels lack the descriptive depth and context clues that textual labels provide, crucial for LLMs trained on natural language text.\\
5) \textbf{Omitting the unmentioned targets may not be a good choice}: While the \textit{OU} option, which excludes unmentioned aspects, might seem to simplify outputs, it also introduces format inconsistency. This lack of uniformity forces the model to adapt to varied aspect mentions per sample, increasing task complexity with dynamic adjustment of the output format. Instead, the \textit{PU} option keeps a consistent output format by adding placeholders, perhaps making LLMs easier to learn. Additional analysis shows that the aspects with a higher degree of unmentioning suffer greater underperformance with \textit{OU} compared to \textit{PU}, see Appendix \ref{app:ou_pu}.

\noindent\textbf{\textit{Conclusions for Reasoning Options:}}\\
1) \textbf{Subtle impact of CoT on ID, while significant on OOD tasks}: CoT design marginally affects ID tasks but markedly improves OOD performance. This contrast highlights CoT's role in enhancing model reasoning and adaptability in unfamiliar contexts, underpinning its value for generalization.\\
2) \textbf{"Think before predict" beats "predict then explain"}: When the reasoning step is placed after predicting, like the \textit{R-CoT} method, the performance does not match that of the standard CoT approach. However, \textit{R-CoT} can still outperform \textit{No-CoT} in many cases, suggesting that a single reasoning component is also beneficial.

\subsubsection{Format Adherence Performance}

\begin{figure*}
    \centering
    \includegraphics[width=1\linewidth]{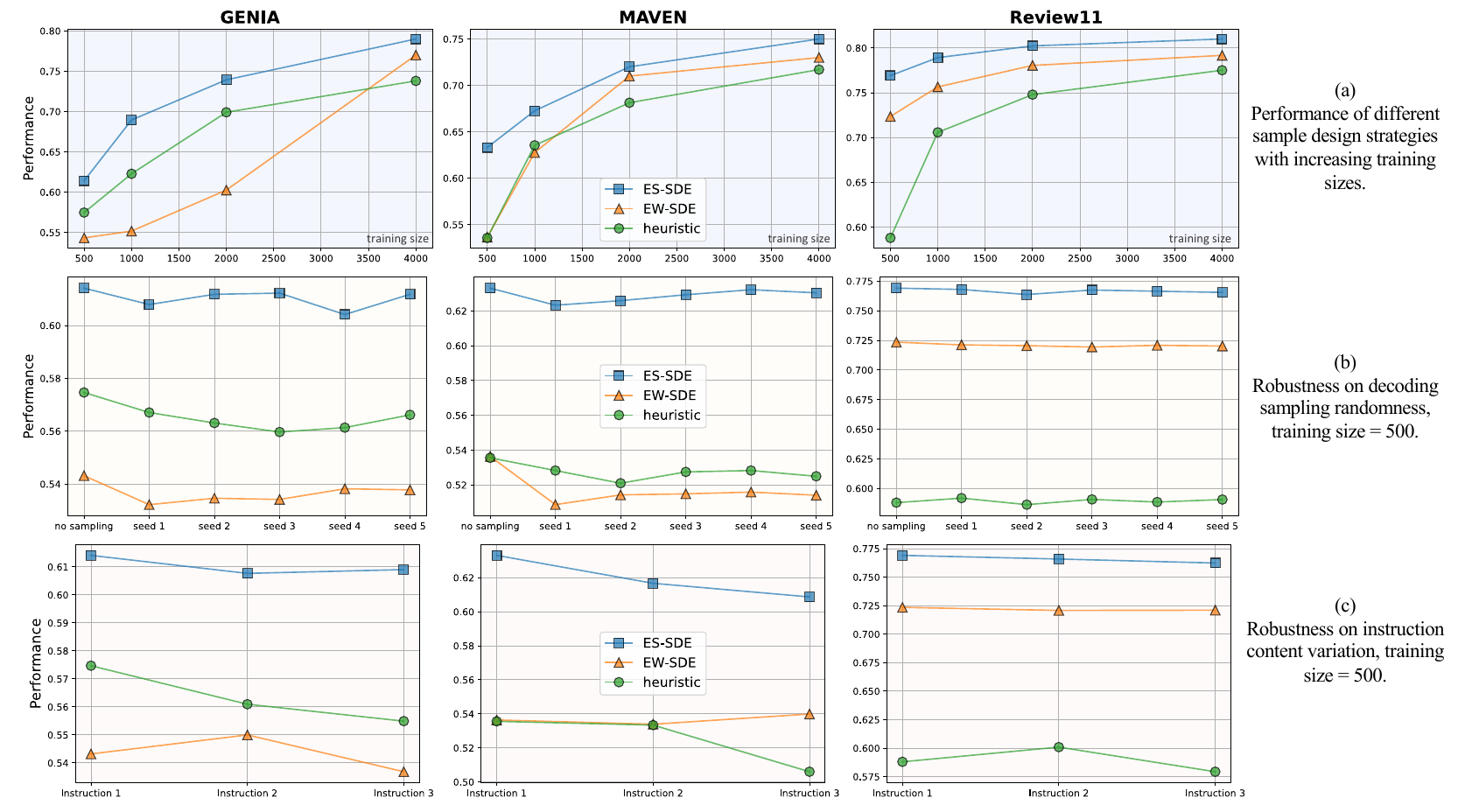}
    \caption{Comparison of different sample design strategies.}
    \label{fig:3downstream3exp}
\end{figure*}

Figure \ref{fig:error-rate} presents the results of the format adherence performances for Chat-LLMs, from which we find that: 1) While the \textit{Inst-first} method improves sentiment analysis, it shows less stability in format adherence, especially in OOD scenarios, indicating that leading with instructions might increase format errors with unfamiliar content;
2) Structured design options lead to better format adherence abilities: A noticeable trend is that structured outputs, especially in the order \textit{JSON} > \textit{Lines} > \textit{Natural}, have lower format error rates. \textit{JSON} format, in particular, demonstrates strong adherence to the correct structure, highlighting a balance between output complexity and precision;
3) \textit{MI}, \textit{NumLabel} and \textit{CoT} options can be quite unstable for certain LLMs, while other options are generally consistent across different models. In applications where stability is vital, these unstable options should be taken seriously;
4) Though improving the understanding or reasoning performances, CoT design puts LLMs at a higher risk of parsing failure for customized downstream tasks, underlining a trade-off for this option.

Considering LLMs' format adherence alongside the understanding abilities is crucial for specialized downstream applications, suggesting a need for a balanced approach in industrial scenarios.

\section{Experiments \uppercase\expandafter{\romannumeral2}: An Robust Integrated SDE Strategy}\label{sec:Exp-2}

Based on the experimental evidence from the previous section, we propose an \textbf{empirically strong SDE strategy} (termed as \textbf{ES-SDE}) using the well-performing options: a combination of \textit{Inst-first}, \textit{No-MI} input designs and \textit{Lines}, \textit{PU}, \textit{TxtLabel} output designs. We don't use the \textit{CoT} design because of its high annotation cost and relatively unstable output. In this section, we conduct comprehensive experiments to validate its effectiveness across different downstream tasks, as well as the robustness against perturbations in instructions or generation.

\subsection{Settings}
\textbf{Tasks and datasets.} To evaluate the effectiveness of ES-SDE, we conduct experiments on three challenging downstream tasks:\\
$\bullet$ \textbf{GENIA} \cite{ohta2002_NER_genia}. A nested named entity recognition (Nested-NER) dataset in the molecular biology domain, where ChatGPT (GPT-3.5) only achieves an F1 score of 50.89\%, using 5-shot CoT reasoning \cite{han2023_IE_evaluation}.\\
$\bullet$ \textbf{MAVEN} \cite{wang2020_EE_maven}. A general domain event detection (ED) dataset. \citet{han2023_IE_evaluation} demonstrate that the performance of ChatGPT in ED tasks falls below expectations. We use the top-10 event types in our experiments.\\
$\bullet$ \textbf{Review11}. This is our self-collected Chinese MASA dataset that involves 11 aspects, more complicated than the MASA tasks in Section \ref{sec:Exp-1}.

\noindent \textbf{Baselines.} As a comparison to \textbf{ES-SDE}, we also propose an \textbf{empirically weak SDE strategy} (\textbf{EW-SDE}), combining \textit{Inst-last}, \textit{Natural}, and \textit{OU}, while keeping other options the same. We naturally hypothesize that EW-SDE should be weaker than ES-SDE.  Note that ES-SDE and EW-SDE are both evidence-based strategies according to the previous empirical results, therefore, we also set up a \textbf{heuristic}-based baseline, referring to the prompt designs from the study of \citet{han2023_IE_evaluation}, which are similar to a combination of \textit{Inst-first} and \textit{OU} options, with a "lines-of-list" output format. Examples of these strategies see Appendix \ref{fig:examples-of-strategies-3new}. 

\noindent\textbf{\textit{Models.}} For a more generalized evaluation, we utilize two new LLMs, instead of those used in Section \ref{sec:Exp-1}. Considering the task language, the \textit{llama2-7b-chat} \cite{touvron2023llama} is used for GENIA and MAVEN and \textit{qwen1.5-4b-chat} \cite{qwen}, a very latest LLM, is used for Review11. The training details are the same as Section \ref{sec:Exp-1}.

\subsection{Results}
Figure \ref{fig:3downstream3exp} reports the comparison between different sample design strategies, from different perspectives. Soft-match F1 scores \cite{han2023_IE_evaluation} are reported for GENIA and MAVEN, and $\kappa$ reported for Review 11. More detailed results see Appendix \ref{app:experiment2}.
Several key conclusions can be observed:

1) \textbf{\textit{ES-SDE maintains advantages across tasks and training sizes.}} Figure \ref{fig:3downstream3exp}-(a) demonstrates a consistent trend that \textbf{ES-SDE} keeps its advantage as the training size increases from $500$ to $4,000$. Notably, $500$ ES-SDE samples worth $\sim2,000$ EW-SDE and heuristic samples in GENIA and Review11 tasks, indicating the high quality of ES-SDE samples.
2) \textbf{\textit{Stable on decoding randomness.}}
By default, the model employs a greedy decoding strategy (no sampling). Figure \ref{fig:3downstream3exp}-(b) shows the results when activating decoding sampling with varying random seeds. \textbf{ES-SDE} maintains exceptional stability across different seeds on three tasks. The adoption of decoding sampling tends to diminish the performances of both SW-SDE and heuristic strategies for GENIA and MAVEN, while ES-SDE gives stable performances. 
3) \textbf{\textit{Robust to instruction variation.}}
 For instructions about a specific task, we have various ways of expressing the same idea. Therefore, we validate the sensitivity of different strategies to different formulations of the instruction, by changing the common content to other formulations (examples in Appendix \ref{fig:new_instructions}). As shown in Figure \ref{fig:3downstream3exp}-(c), ES-SDE keeps its edge in different variations, showing its robustness to instruction content.

Overall, \textbf{ES-SDE} represents a reliable and potent approach for the DT of LLMs, illustrating that—through a careful SDE process, LLMs can achieve much higher performances in downstream tasks. Note that ES-SDE may not be the best strategy for all tasks. A detailed investigation into SDE across a broader spectrum of tasks and models could yield even more effective strategies.

\section{Can PE guide SDE? An Additional Analysis}
\begin{figure}
    \centering
    \includegraphics[width=1\linewidth]{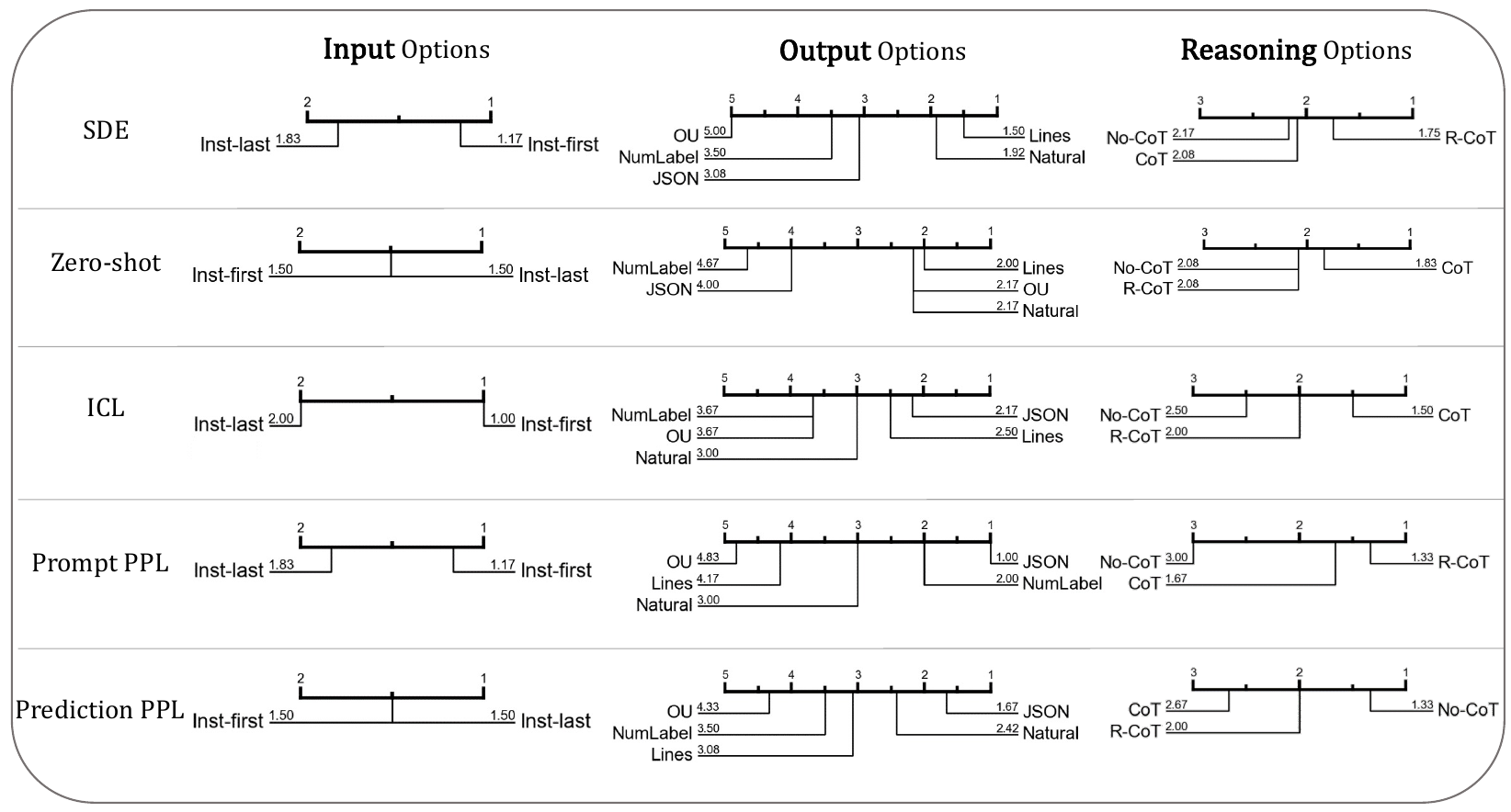}
    \caption{Average rankings of the DT performances of SDE options and zero-shot/ICL/PPL rankings of their corresponding prompts. Results based on the MASA ID tasks across 6 LLMs.}
    \label{fig:can_PE_guide_SDE}
\end{figure}

Prompts are the key to understand models' innate qualities and capabilities. A good PE method often indicates some patterns that a LLM is more familiar with or excels in. A natural question is: \textit{can PE guide SDE?} 
To answer this question, we craft zero-shot and ICL prompts according to different SDE options to evaluate their PE performances. Figure \ref{fig:can_PE_guide_SDE} reports the average rankings of SDE options and their corresponding prompts in the MASA ID tasks. Detailed results for each task see Appendix \ref{app:pe_analysis}.

Our analysis revealed some consistent patterns: \textit{Inst-first} is an effective choice for both PE and SDE; \textit{CoT} improves performances for both PE and SDE evaluations. 
However, there are also many counter-intuitive findings. For example, the \textit{OU} option consistently harms DT performances according to our previous experiments, however, its corresponding prompts results in notably better zero-shot or ICL results for certain LLMs; Similarly, while the \textit{Natural} option outperforms the \textit{Lines} approach for base models in SDE, the reverse is true in zero-shot or ICL evaluations for models like c-llama2-base and intern-base. 
\citet{gonen2023prompt_ppl} showed through a wide range of tasks that the lower that lower perplexity (PPL) generally leads to better prompt designs. Inspired by this, we also conduct PPL analysis on the ICL prompts/predictions corresponding to each SDE options. Interestingly, \textit{OU}-like prompt gives the highest averaged PPL scores across all options, which seems to be contradictory that \textit{OU} brings better zero-shot or ICL results. The \textit{JSON} format surprisingly achieves rather low PPL scores, however its SDE performances are worse than \textit{Lines}.

These findings highlight a complex landscape where \textbf{prompt design patterns do not always align with SDE effectiveness}, underscoring the nuanced relationship between PE and SDE.

\section{Conclusion \& Future Work}
In this study, we introduce SDE as an effective method to enhance the downstream-tuning performances of LLMs. Through comprehensive ID and OOD experiments involving six LLMs, we demonstrate the effects of various sample design strategies, uncovering some interesting patterns that are consistent across different LLMs. Building on these findings, we develop the ES-SDE approach, which integrates the most effective options. Our experiments on three new tasks with two additional LLMs consistently show ES-SDE's superiority over baseline methods. Further analysis of the relationship between PE and SDE suggests that effective prompt designs do not necessarily translate to successful sample designs. This observation opens up avenues for more detailed investigations into the mechanisms of SDE in future research.

\section{Limitations}
This research follows a two-step experimental approach. In the first step, we investigate the impact of each SDE option, the results are then used as evidence for the second step—proposing an empirically strong SDE combination strategy. As an empirical study, this research is subject to certain limitations:
\begin{itemize}
    \item[1.] While we demonstrate that the experimental findings from the first phase are extendable to different downstream tasks, the applicability to other untested scenarios remains uncertain. For instance, although the \textit{Lines} output design outperforms the \textit{JSON} format in our current experiments, it is unclear if this advantage persists in more complex tasks with intricate structures. Future research will address these more challenging contexts;
    \item[2.] With the rapid pace of advancements in LLMs, new and more sophisticated models are being introduced frequently. The models we used in our study were among the best open-source options available at the start of our research but have since been surpassed by newer releases. Although we assessed a total of 8 LLMs, including both base and chat variants, there remains a possibility that our findings may not be universally applicable to other models;
    \item[3. ] Combining different SDE options poses significant challenges, particularly without prior validation experiments such as those described in Section \ref{sec:Exp-1}. The challenges are twofold. Firstly, unlike typical hyperparameters like learning rate or network layers, choosing different SDE options alters the training data itself, rendering traditional hyperparameter-tuning techniques such as Bayesian Optimization \cite{snoek2012_bayesian_opt} less practical. Secondly, evaluating LLMs on downstream tasks is both resource-intensive and costly, due to the need for customized task metrics, parsing rules, and high model inference costs. Therefore, developing a more efficient framework for SDE studies is a critical objective for future research.  
\end{itemize}

\bibliography{SDE}

\appendix

\section{Appendix}\label{sec:appendix}

\subsection{Metrics for MASA}\label{app:metrics}

\textbf{Weighted Kappa.} 
Considering the imbalance of different aspects and the ordinal nature of labels, weighted agreement measures are proved to be more effective than traditional metrics \cite{BENDAVID2008825,GALAR20111761,grandini2020metrics}. Thus we adopt Weighted Kappa \citep{cohen1968weighted, YILMAZ2023110020} as the measure of classification effect, which is an extension of Cohen's Kappa \citep{cohen1960coefficient}. Weighted Kappa $\kappa$ is defined as $\kappa=\frac{P_o-P_e}{1-P_e}$, which measures a model's performance by considering how much better it performs than random guessing. Here, $P_o=\sum_{i,j=1}^{R} {w_{ij}p_{ij}}$ and $P_e=\sum_{i,j=1}^{R} {w_{ij}p_{i.}p_{.j}}$. The probabilities $p_{ij},p_{i.},p_{.j}$ are values or accumulated values from the classification confusion matrix. The weighting factor, $w_{ij}$, enables a nuanced assessment of different error degrees. For example, classifying "positive" as "negative" is more detrimental than classifying "positive" as "neutral," hence a higher penalty should be imposed on the former. Based on the feedback from enterprises in practical applications, we define the weight matrix without loss of generality as Table \ref{table:weighted_kappa_matrix}. \\
\begin{table}[h]
\small
\centering
\renewcommand{\arraystretch}{1.5}
\setlength\tabcolsep{2pt}
\begin{tabular}{c|cccc}
&Pre-Pos&Pre-Neu&Pre-Neg&Pre-Unm\\
\hline
Label-Pos&1& 1/2  &0& 1/2\\
Label-Neu& 2/3&1&2/3&2/3\\
Label-Neg&0&1/2&1&1/2\\
Label-Unm&1/2&2/3&1/2&1\\
\end{tabular}
\caption{Weight matrix for calculating weighted Kappa.}
\label{table:weighted_kappa_matrix}
\end{table}
\begin{table*}[h]
\small
\centering
\setlength\tabcolsep{6pt}
\begin{tabular}{ll|cccc|cccc|cccc}
\toprule[1.2pt]
   &        & \multicolumn{4}{c|}{TrainSet (size=500)}     & \multicolumn{4}{c|}{TrainSet (size=1000)}    & \multicolumn{4}{c}{TestSet}           \\
\midrule
 &  & Pos & Neu & Neg & Unm & Pos & Neu & Neg & Unm & Pos & Neu & Neg & Unm \\
\midrule
\multirow{6}{*}{\textbf{D1}} & F   & 65.20 & 15.00 & 18.80 & 1.00  & 66.60 & 13.70 & 18.30 & 1.40  & 66.01 & 12.23 & 20.12 & 1.64  \\
   & B   & 22.20 & 4.20  & 8.20  & 65.40 & 23.50 & 3.60  & 7.20  & 65.70 & 21.50 & 3.15  & 6.29  & 69.07 \\
   & P   & 33.40 & 13.00 & 15.60 & 38.00 & 35.60 & 10.70 & 15.80 & 37.90 & 36.64 & 10.24 & 13.97 & 39.15 \\
   & H   & 14.80 & 1.20  & 6.00 & 78.00 & 17.10 & 1.00  & 5.50  & 76.40 & 16.12 & 0.82  & 5.58 & 77.48 \\
   & SA & 48.80 & 3.60  & 14.00 & 33.60 & 47.90 & 4.10  & 13.60 & 34.40 & 42.73 & 3.46  & 13.87 & 39.94 \\
   & PC & 4.40  & 0.60 & 1.40  & 93.60 & 4.80  & 0.30  & 1.90  & 93.00 & 3.93  & 0.34 & 1.56 & 94.18 \\
\midrule
\multirow{5}{*}{\textbf{D2}} & TC & 52.40 & 13.20 & 7.60  & 26.80 & 53.10 & 13.20 & 8.10  & 25.60 & 48.56 & 12.84 & 7.03  & 31.57 \\
   & Q & 18.80 & 8.20 & 11.20 & 61.80 & 17.90 & 10.10 & 11.00 & 61.00 & 14.67 & 10.00 & 10.44 & 64.89 \\
   & SS & 16.80 & 3.60 & 8.20 & 71.40 & 15.70 & 3.80  & 8.90  & 71.60 & 14.86 & 3.15 & 8.58  & 73.41 \\
   & D   & 46.00 & 8.20  & 4.20  & 41.60 & 48.50 & 8.10  & 4.30  & 39.10 & 43.10 & 7.68  & 5.28  & 43.93 \\
   & N   & 1.00 & 1.40 & 2.80 & 94.80 & 1.40 & 1.30 & 3.40 & 93.90 & 2.10 & 1.08 & 3.36 & 93.46 \\
\bottomrule[1.2pt]
\end{tabular}
\caption{Label distribution(\%) in various aspects of train set and test set. \textbf{D1} contains annotations for 6 aspects—food (F), beverage (B), price (P), hygiene (H), staff attitude (SA), and parking convenience (PC); \textbf{D2} contains annotations for 5 different aspects—traffic convenience (TC), queuing (Q), serving speed (SS), decoration (D), and noise (N). We use 'Pos', ‘Neu’, 'Neg', ‘Unm’ to represent Positive, Neutral, Negative and Unmentioned labels, respectively.}
\label{table:label-distribution}
\end{table*}

\begin{figure*}[hb]
    \centering
    \includegraphics[width=1\linewidth]{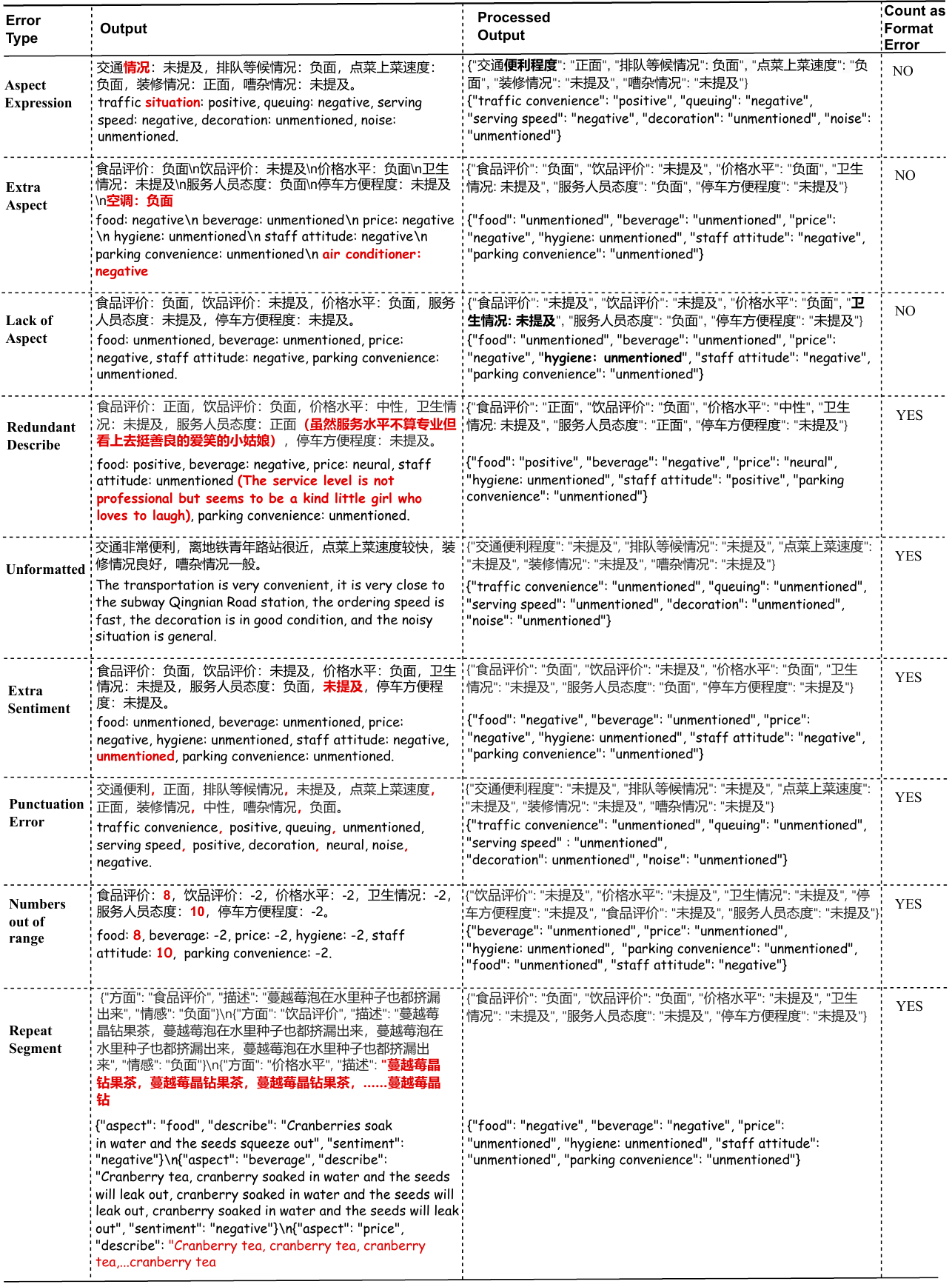}
    \caption{Examples of format error types and how they are processed.}
    \label{fig:format_errors}
\end{figure*}

\noindent\textbf{Format adherence.} Format adherence not only ensures that outputs from the model can be reliably parsed and utilized in practical applications, but also reflects the model's ability to understand the context and the nuances of different instructions. 
We set up parsers according to the prescribed formats of different designs, then we calculate the ratio of predictions that cannot be successfully parsed with our output parser. 
Considering the inherently uncertainty nature of generative language models, 
we relaxed the format such as the expression of aspects and sentiments.
Meanwhile, in order to compare the content correctness between designs more fairly, 
for some cases such as common punctuation errors, we will correct it into the required format when calculating the Kappa. Figure \ref{fig:format_errors} shows a variety of representative format error types and how they are processed by the parsers we design.

\subsection{Datasets and Training Settings}\label{app:datasets}
Table \ref{table:label-distribution} shows the label distribution of each aspect for two domains \textbf{D1} and \textbf{D2}, where we can see the distributions are highly unbalanced.

The training setup was as follows: learning rate
set to 1e-4, batch size of 4, LoRA rank of 8
LoRA alpha of 32, LoRA dropout of 0.1. 

\subsection{Sample Design Examples}\label{app:examples}
\begin{figure*}[t]
    \centering   \includegraphics[width=\textwidth]{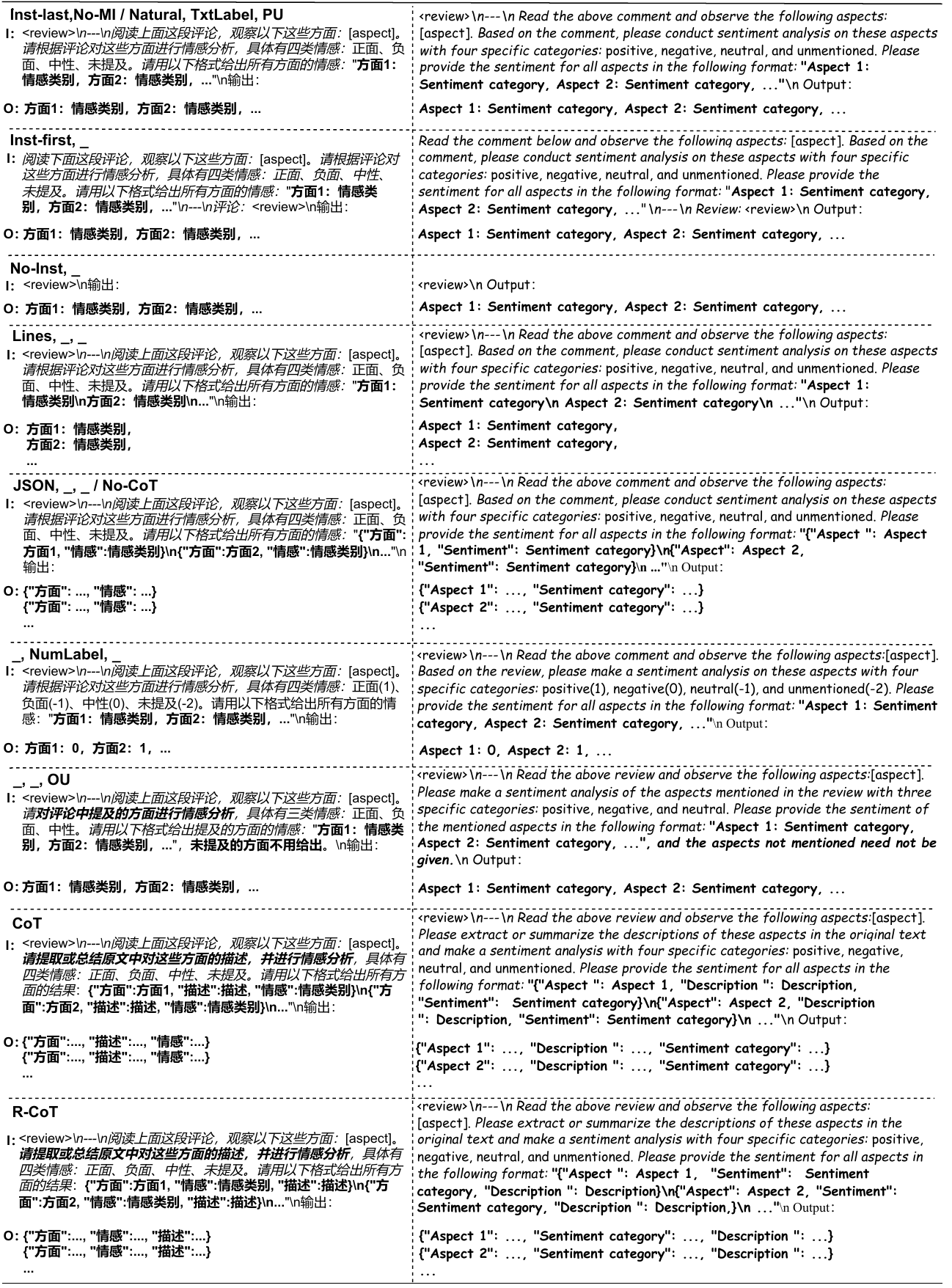}
    \caption{Examples of different sample designs on the MASA task.}
    \label{fig:examples-of-strategies}
\end{figure*}

Figure \ref{fig:examples-of-strategies} shows a detailed example of our sample designs on MASA tasks.

\subsection{Detailed Evaluations of Each SDE Option}\label{app:experiment1}

The detailed results of in-domain (ID) and out-of-domain (OOD) evaluations on the MASA task of different SDE options across six LLMs are shown in Table \ref{fig:LLaMA2-Chat_result} to Table \ref{fig:Baichuan2-Base_result}, including both the sentiment analysis performances ($\kappa$) and the format adherence performances (format error rate). An averaged results of training size 500 and 1000 of ID and OOD scenarios are visualized in Figure \ref{fig:big-results-figure}.

\begin{table*}
\centering
\small
\renewcommand{\arraystretch}{0.9}
\setlength\tabcolsep{4pt}
\begin{tabular}{ll|cc|cc|cc|cc}
\toprule[1.2pt]
   \multicolumn{2}{c|}{model: \textbf{c-llama2-chat}} &\multicolumn{4}{c|}{\textbf{Weighted Kappa $\kappa$}} &\multicolumn{4}{c}{\textbf{\# Wrong format} {\scriptsize(7969 test samples in total)}}  \\
\midrule
  \multicolumn{2}{c|}{train\_size=500} & D1→D1 & D2→D2 & D1→D2 & D2→D1  & D1→D1 & D2→D2 & D1→D2 & D2→D1    \\
\midrule

\multirow{4}{*}{Input} &
  Inst-last, No-MI &
  .8091 &
  .6882 &
  .5243 &
  .7217 &
  0 & 0 & 2 &2 \\
 
 & Inst-first, \_ & .8136 & .7079 & .5124 & .7223 & 0  & 0  & 9   & 15   \\
 
 &No-inst, \_ &
  .7757 &
  .6626 &
  \textbackslash{} &
  \textbackslash{} &
  20 &
  1 &
  \textbackslash{} &
  \textbackslash{} \\
 
 & \_, MI        & .6187 & .6187 & .4806 & .2756 & 1  & 0  & 0   & 1079 \\
\midrule

\multirow{4}{*}{Output} &
  Natural, TxtLabel, PU &
  .8091 &
  .6882 &
  .5243 &
  .7217 &
  0 &
  0 &
  2 &
  2 \\
 
 & Lines, \_, \_     & .8083 & .6969 & .5068 & .7447 & 0  & 0  & 0   & 0    \\
 
 & JSON, \_, \_      & .8086 & .6952 & .4905 & .7354 & 0  & 0  & 0   & 0    \\
 
 & \_, NumLabel, \_ & .7697 & .6373 & .4221 & .6723 & 3  & 1  & 0   & 1260 \\
 
 & \_, \_, OU        & .7934 & .6005 & .5282 & .6203 & 0  & 0  & 87  & 0    \\
\midrule

\multirow{3}{*}{Reasoning} &No-CoT &
  .7934 &
  .6005 &
  .5282 &
  .6203 &
  0 &
  0 &
  87 &
  0 \\
  
 & CoT       & .7928 & .6873 & .5249 & .7085 & 56 & 65 & 36  & 282  \\
 
 & R-CoT   & .8074 & .6752 & .4726 & .7297 & 93 & 65 & 141 & 263 \\
 
\midrule
  \multicolumn{2}{c|}{train\_size=1000} & D1→D1 & D2→D2 & D1→D2 & D2→D1  & D1→D1 & D2→D2 & D1→D2 & D2→D1    \\
\midrule

\multirow{4}{*}{Input} &Inst-last, No-MI  & 0.8256 & 0.7110 & 0.5518 & 0.7312 & 0  & 0  & 0   & 3   \\

&Inst-first, \_ & 0.8236 & 0.7090 & 0.5483 & 0.7264 & 0  & 0  & 5   & 1   \\

&No-inst, \_ & 0.8003 & 0.6920 & \textbackslash{} & \textbackslash{} & 6 & 4 & \textbackslash{} & \textbackslash{} \\

& \_, MI        & 0.8113 & 0.6700 & 0.5095 & 0.5182 & 0  & 0  & 0   & 728 \\
\midrule
          
\multirow{5}{*}{Output} &Natural, TxtLabel, PU   & 0.7916 & 0.7253 & 0.5303 & 0.7356 & 0 & 0  & 0 & 3 \\

&Lines, \_, \_     & 0.8259 & 0.7118 & 0.5560 & 0.7452 & 0  & 0  & 0   & 0   \\

&JSON, \_, \_      & 0.8249 & 0.7094 & 0.5488 & 0.7432 & 0  & 0  & 0   & 0   \\

& \_, NumLabel, \_ & 0.7624 & 0.6604 & 0.4210 & 0.6840 & 2  & 2  & 0   & 765 \\

& \_, \_, OU        & 0.8172 & 0.7125 & 0.5511 & 0.6746 & 0  & 0  & 493 & 1   \\
\midrule
          
\multirow{3}{*}{Reasoning} &No-CoT    & 0.8018 & 0.7175 & 0.5332 & 0.7323 & 0  & 0  & 493 & 1   \\

&CoT       & 0.8111 & 0.7111 & 0.5354 & 0.7311 & 59 & 24 & 30  & 253 \\

&R-CoT   & 0.8214 & 0.7137 & 0.5085 & 0.7532 & 51 & 25 & 75  & 115\\

\bottomrule[1.2pt]
\end{tabular}
\caption{MASA evaluations of each SDE option for model \textbf{c-llama2-chat}. The first method in each group is the group baseline. "\_" means keeping the same option with the group baseline.}
\label{fig:LLaMA2-Chat_result}
\end{table*}

\begin{table*}
\centering
\small
\renewcommand{\arraystretch}{0.9}
\setlength\tabcolsep{4pt}
\begin{tabular}{ll|cc|cc|cc|cc}
\toprule[1.2pt]
   \multicolumn{2}{c|}{model: \textbf{c-llama2-base}} &\multicolumn{4}{c|}{\textbf{Weighted Kappa $\kappa$}} &\multicolumn{4}{c}{\textbf{\# Wrong format} {\scriptsize(7969 test samples in total)}}  \\
\midrule
  \multicolumn{2}{c|}{train\_size=500} & D1→D1 & D2→D2 & D1→D2 & D2→D1  & D1→D1 & D2→D2 & D1→D2 & D2→D1    \\
\midrule

\multirow{4}{*}{Input} & Inst-last, No-MI & 0.8067 & 0.6801 & 0.5246  & 0.7000  & 0 & 0 & 6 & 98 \\

&Inst-first, \_ & 0.8092 & 0.6921 & 0.5575 & 0.6794 & 0  & 0  & 34  & 3    \\

&No-inst, \_      & 0.7762 & 0.6511 & \textbackslash{} & \textbackslash{} & 0 & 1                        & \textbackslash{} & \textbackslash{}          \\

& \_, MI        & 0.7778 & 0.5024 & 0.4946 & 0.4184 & 2  & 0  & 118 & 0    \\
\midrule
          
\multirow{5}{*}{Output} &Natural, TxtLabel, PU  & 0.8067 & 0.6801 & 0.5246 & 0.7000  & 0 & 0 & 6 &  98 \\

&Lines, \_, \_     & 0.8066 & 0.6410 & 0.5128 & 0.6622 & 0  & 0  & 19  & 0    \\

&JSON, \_, \_      & 0.8010 & 0.6242 & 0.5170 & 0.6287 & 0  & 0  & 0   & 0    \\

& \_, NumLabel, \_ & 0.7728 & 0.5949 & 0.5155 & 0.6296 & 14 & 1  & 26  & 356  \\

& \_, \_, OU        & 0.7746 & 0.5012 & 0.4199 & 0.5711 & 0  & 3  & 300 & 7    \\
\midrule
          
\multirow{3}{*}{Reasoning} &No-CoT    & 0.8010 & 0.6242 & 0.5170 & 0.6287 & 0  & 0  & 0   & 0    \\

&CoT       & 0.7789 & 0.6652 & 0.4649 & 0.6974 & 83 & 82 & 33  & 226  \\

&R-CoT   & 0.8019 & 0.6428 & 0.4657 & 0.4199 & 88 & 11 & 87  & 1823\\

\midrule[1.2pt]
  \multicolumn{2}{c|}{train\_size=1000} & D1→D1 & D2→D2 & D1→D2 & D2→D1  & D1→D1 & D2→D2 & D1→D2 & D2→D1    \\
\midrule

\multirow{4}{*}{Input}& Inst-last, No-MI  & 0.8237 & 0.7011 & 0.6010 & 0.7197 & 0  & 0  & 3   & 177 \\

 & Inst-first, \_ & 0.8231 & 0.7068 & 0.6069 & 0.6956 & 0  & 2  & 16  & 28  \\
 
 & No-inst, \_ & 0.7957 & 0.6882 & \textbackslash{} & \textbackslash{} & 2 & 2 & \textbackslash{} & \textbackslash{} \\
 
 & \_, MI        & 0.8048 & 0.6174 & 0.5306 & 0.6390 & 0  & 3  & 139 & 6   \\
\midrule

\multirow{5}{*}{Output}& Natural, TxtLabel, PU   & 0.8237 & 0.7011 & 0.6010 & 0.7197 & 0  & 0  & 3   & 177 \\
 
 & Lines, \_, \_     & 0.8205 & 0.6947 & 0.5900 & 0.6963 & 0  & 0  & 10  & 0   \\
 
 & JSON, \_, \_      & 0.8212 & 0.6857 & 0.5649 & 0.6875 & 0  & 0  & 0   & 0   \\
 
 & \_, NumLabel, \_ & 0.7619 & 0.6536 & 0.4804 & 0.6709 & 1  & 2  & 0   & 584 \\
 
 & \_, \_, OU        & 0.8179 & 0.6774 & 0.5034 & 0.6277 & 0  & 5  & 64  & 29  \\
\midrule

\multirow{3}{*}{Reasoning}& No-CoT    & 0.8212 & 0.6857 & 0.5649 & 0.6875 & 0  & 0  & 0   & 0   \\
 
 & CoT       & 0.8026 & 0.6979 & 0.5519 & 0.7159 & 70 & 31 & 16  & 125 \\
 
 & R-CoT   & 0.8195 & 0.7034 & 0.5368 & 0.6454 & 46 & 14 & 24  & 666\\

\bottomrule[1.2pt]

\end{tabular}
\caption{MASA evaluations of each SDE option for model \textbf{c-llama2-base}. Definition of "\_" see Table \ref{fig:LLaMA2-Chat_result}.}
\label{fig:LLaMA2-Base_result}
\end{table*}

\begin{table*}
\centering
\small
\renewcommand{\arraystretch}{0.9}
\setlength\tabcolsep{4pt}
\begin{tabular}{ll|cc|cc|cc|cc}
\toprule[1.2pt]
   \multicolumn{2}{c|}{model: \textbf{intern-chat}} &\multicolumn{4}{c|}{\textbf{Weighted Kappa $\kappa$}} &\multicolumn{4}{c}{\textbf{\# Wrong format} {\scriptsize(7969 test samples in total)}}  \\
\midrule
  \multicolumn{2}{c|}{train\_size=500} & D1→D1 & D2→D2 & D1→D2 & D2→D1  & D1→D1 & D2→D2 & D1→D2 & D2→D1    \\
\midrule

  \multirow{4}{*}{Input}&Inst-last, No-MI  & 0.7774 & 0.6278 & 0.3947 & 0.6707  & 0  & 0  & 0 & 11 \\

  &Inst-first, \_ & 0.8035 & 0.6609 &0.3949  & 0.7090 & 4  & 2  & 13 & 304\\

  &T2L & 0.7862 & 0.5963 & \textbackslash{} & \textbackslash{} & 10 & 7  & \textbackslash{} & \textbackslash{} \\

  & \_, MI  & 0.7463 & 0.5178 & 0.3153 & 0.5363& 0  & 0  & 0 & 395 \\
\midrule

  \multirow{5}{*}{Output}& Natural, TxtLabel, PU & 0.7774 & 0.6278 & 0.3947 & 0.6707  & 0 & 0 & 0 & 11 \\

  &Lines, \_, \_ & 0.7827 & 0.6261 & 0.4032 & 0.6799 & 0  & 1  & 1 & 1 \\

  &JSON, \_, \_& 0.7713 & 0.5966 & 0.3965  & 0.6129 & 0  & 0  & 0  & 2 \\
       
  & \_, NumLabel, \_& 0.7765 & 0.6261 & 0.4165 & 0.6926 & 0  & 0  & 3 & 23  \\
       
  & \_, \_, OU & 0.7520 & 0.4888 & 0.4029 & 0.6221 & 0  & 1  & 16 & 7 \\
\midrule

  \multirow{3}{*}{Reasoning}&No-CoT& 0.7713 & 0.5966 & 0.3965 & 0.6129 & 0  & 0  & 0 & 2 \\

  & CoT & 0.7666 & 0.6401 & 0.4843 & 0.6797 & 43 & 19 & 30  & 121 \\
       
  &R-CoT& 0.7764 & 0.6124 & 0.3892 & 0.6648 & 44 & 23 & 23 & 72 \\

\midrule[1.2pt]

  \multicolumn{2}{c|}{train\_size=1000} & D1→D1 & D2→D2 & D1→D2 & D2→D1  & D1→D1 & D2→D2 & D1→D2 & D2→D1    \\
\midrule

  \multirow{4}{*}{Input}  &Inst-last, No-MI & 0.8049 & 0.6793 & 0.4330 & 0.6982 & 0  & 0 & 0 & 0 \\
  
  &Inst-first, \_ & 0.8173 & 0.7125 & 0.4640 & 0.7343 & 0  & 1 & 6 & 259 \\
  
  &No-inst, \_& 0.8139 & 0.6811 & \textbackslash{} & \textbackslash{} & 8  & 5 & \textbackslash{} & \textbackslash{} \\
  & \_, MI & 0.7819 & 0.6256 & 0.3332 & 0.6520 & 1  & 0 & 8  & 29 \\
\midrule

  \multirow{5}{*}{Output} & Natural, TxtLabel, PU& 0.8049 & 0.6793 & 0.4330 & 0.6982 & 0  & 0 & 0 & 0  \\
  
  &Lines, \_, \_ & 0.8060 & 0.6797 & 0.4498 & 0.7038 & 0  & 1 & 0  & 1 \\
  & JSON, \_, \_ & 0.8021 & 0.6649 & 0.4661 & 0.6647 & 0  & 0 & 0 & 0 \\

  & \_, NumLabel, \_& 0.8081 & 0.6764 & 0.4393 & 0.7286 & 0 & 0 & 3 & 3\\
  
  & \_, \_, OU& 0.8008 & 0.6369 & 0.4374 & 0.6694 & 0  & 0 & 33 & 1 \\
\midrule

  \multirow{3}{*}{Reasoning} &No-CoT & 0.8021 & 0.6649 & 0.4661 & 0.6647 & 0  & 0 & 0 & 0 \\
  
  & CoT & 0.7981 & 0.6966 & 0.5190 & 0.7098 & 36 & 7 & 10 & 132 \\
  
  &R-CoT & 0.8043 & 0.6709 & 0.3994 & 0.7195 & 50 & 4 & 19 & 42\\
\bottomrule[1.2pt]

\end{tabular}
\caption{MASA evaluations of each SDE option for model \textbf{intern-chat}. Definition of "\_" see Table \ref{fig:LLaMA2-Chat_result}.}
\label{fig:Internlm-Chat_result}
\end{table*}
\begin{table*}
\centering
\small
\renewcommand{\arraystretch}{0.9}
\setlength\tabcolsep{4pt}
\begin{tabular}{ll|cc|cc|cc|cc}
\toprule[1.2pt]
  \multicolumn{2}{c|}{model: \textbf{intern-base}} &\multicolumn{4}{c|}{\textbf{Weighted Kappa $\kappa$}} &\multicolumn{4}{c}{\textbf{\# Wrong format} {\scriptsize(7969 test samples in total)}}  \\
\midrule
  \multicolumn{2}{c|}{train\_size=500} & D1→D1 & D2→D2 & D1→D2 & D2→D1  & D1→D1 & D2→D2 & D1→D2 & D2→D1    \\
\midrule

 \multirow{4}{*}{Input}  &Inst-last, No-MI& 0.7849 & 0.6465 & 0.4898 & 0.6129 & 0   & 1  & 1 & 0 \\
 
  &Inst-first, \_  & 0.7955 & 0.6472 & 0.4947 & 0.7006  & 3   & 8  & 18  & 221 \\
                              
  &No-inst, \_& 0.7936 & 0.6119 & \textbackslash{} & \textbackslash{} & 11  & 6  & \textbackslash{} & \textbackslash{} \\
  
  & \_, MI& 0.7562 & 0.5029 & 0.3305 & 0.4672  & 0   & 1  & 232 & 447 \\
\midrule

  \multirow{5}{*}{Output} &Natural, TxtLabel, PU & 0.7849 & 0.6465 & 0.4898 & 0.6129 & 0   & 1  & 1 & 0 \\
  
  &Lines, \_, \_ & 0.7873 & 0.6455 & 0.4939 & 0.6365 & 0   & 2  & 4 & 0 \\
  
  &JSON, \_, \_ & 0.7859 & 0.6250 & 0.4727 & 0.6127 & 0   & 0  & 3 & 82 \\
  
  & \_, NumLabel, \_ & 0.7605 & 0.6003 & 0.3861  & 0.6412 & 14  & 3  & 10 & 102 \\
  
  & \_, \_, OU& 0.7275 & 0.5185 & 0.3943& 0.4935& 0   & 4  & 48& 6\\
\midrule

  \multirow{3}{*}{Reasoning} &No-CoT & 0.7859 & 0.6250 & 0.4727 & 0.6127 & 0   & 0  & 3 & 82 \\
  
  & CoT & 0.7621 & 0.6489 & 0.4581  & 0.6388 & 77 & 12 & 2347 & 50\\
  
  &R-CoT& 0.7734 & 0.6342 & 0.3752 & 0.6816 & 141 & 49 & 1496 & 206\\  
\midrule[1.2pt]

  \multicolumn{2}{c|}{train\_size=1000} & D1→D1 & D2→D2 & D1→D2 & D2→D1  & D1→D1 & D2→D2 & D1→D2 & D2→D1    \\
\midrule

  \multirow{4}{*}{Input} &Inst-last, No-MI& 0.8112 & 0.6874 & 0.5216 & 0.7065 & 1  & 0  & 0 & 0 \\
  
  &Inst-first, \_& 0.8167 & 0.6965 & 0.5195 & 0.7544  & 0  & 0  & 5  & 46 \\
  
  &No-inst, \_ & 0.8191 & 0.6963 & \textbackslash{} & \textbackslash{} & 5  & 8  & \textbackslash{} & \textbackslash{} \\
  
  & \_, MI& 0.7937 & 0.6238 & 0.2780 & 0.6492 & 0  & 2  & 383 & 45 \\
\midrule

  \multirow{5}{*}{Output} &Natural, TxtLabel, PU& 0.8112 & 0.6874 & 0.5216 & 0.7065 & 1 & 0 & 0 & 0 \\
  
  &Lines, \_, \_ & 0.8113 & 0.6919 & 0.5060 & 0.7126 & 0  & 0  & 3 & 0 \\
  
  &JSON, \_, \_&0.8076 & 0.6781 & 0.5195 & 0.6817 & 0  & 0  & 3 & 1 \\
  
  & \_, NumLabel, \_& 0.8084 & 0.6776 & 0.4426 & 0.7139 & 3  & 1  & 31 & 20 \\
  
  & \_, \_, OU& 0.8006 & 0.6330 & 0.4587 & 0.6098 & 0  & 1  & 30 & 3 \\
\midrule

  \multirow{3}{*}{Reasoning} &No-CoT & 0.8076 & 0.6781 & 0.5195 & 0.6817 & 0  & 0  & 3 & 1 \\
  
  &CoT& 0.7956 & 0.6874 & 0.5196 & 0.6903 & 34 & 12 & 405 & 56 \\
  
  &R-CoT & 0.8069 & 0.6725 & 0.4890 & 0.7185 & 46 & 11 & 220 & 125\\
\bottomrule[1.2pt]
\end{tabular}
\caption{MASA evaluations of each SDE option for model \textbf{intern-base}. Definition of "\_" see Table \ref{fig:LLaMA2-Chat_result}.}
\label{fig:Internlm-Base_result}
\end{table*}

\begin{table*}
\centering
\small
\renewcommand{\arraystretch}{0.9}
\setlength\tabcolsep{4pt}
\begin{tabular}{ll|cc|cc|cc|cc}
\toprule[1.2pt]
   \multicolumn{2}{c|}{model: \textbf{bc2-chat}} &\multicolumn{4}{c|}{\textbf{Weighted Kappa $\kappa$}} &\multicolumn{4}{c}{\textbf{\# Wrong format} {\scriptsize(7969 test samples in total)}}  \\
\midrule
  \multicolumn{2}{c|}{train\_size=500} & D1→D1 & D2→D2 & D1→D2 & D2→D1  & D1→D1 & D2→D2 & D1→D2 & D2→D1    \\
\midrule

  \multirow{4}{*}{Input}&Inst-last, No-MI& 0.7904 & 0.6544 & 0.4067 & 0.6170 & 8  & 0  & 21 & 10  \\
  
&Inst-first, \_& 0.7958 & 0.6660 & 0.3858 & 0.6739 & 19 & 36 & 12 & 385 \\

&No-inst, \_& 0.7176 & 0.4776 & \textbackslash{} & \textbackslash{} & 23   & 13  & \textbackslash{} & \textbackslash{} \\

& \_, MI& 0.7645 & 0.5636 & 0.3713 & 0.5490 & 0  & 0  & 5  & 16  \\
\midrule

\multirow{5}{*}{Output}&Natural, TxtLabel, PU  & 0.7904 & 0.6544 & 0.4067 & 0.6170 & 8  & 0  & 21 & 10  \\

&Lines, \_, \_ & 0.7869 & 0.6653 & 0.4091 & 0.6344 & 0  & 0  & 9  & 1   \\

&JSON, \_, \_& 0.7927 & 0.6489 & 0.4714 & 0.6196 & 0  & 0  & 1  & 0   \\

& \_, NumLabel, \_& 0.7839 & 0.6401 & 0.3671 & 0.6506 & 5  & 4  & 12 & 17  \\

& \_, \_, OU& 0.7016 & 0.5670 & 0.3599 & 0.3285 & 2  & 81 & 50 & 19  \\
\midrule

\multirow{3}{*}{Reasoning}&No-CoT & 0.7927 & 0.6489 & 0.4714 & 0.6196 & 0  & 0  & 1  & 0   \\

&CoT& 0.7722 & 0.6400 & 0.5006 & 0.6776 & 3641 & 757 & 739 & 3323 \\

&R-CoT & 0.7922 & 0.6535 & 0.4534 & 0.6579 & 107 & 126 & 280 & 563 \\

\midrule[1.2pt]
  \multicolumn{2}{c|}{train\_size=1000} & D1→D1 & D2→D2 & D1→D2 & D2→D1  & D1→D1 & D2→D2 & D1→D2 & D2→D1    \\
\midrule

\multirow{4}{*}{Input}&Inst-last, No-MI& 0.8113 & 0.7060 & 0.4709 & 0.6365 & 0  & 4  & 13 & 18  \\

&Inst-first, \_& 0.8142 & 0.7095 & 0.4733 & 0.6787 & 31 & 12 & 21 & 136 \\

&No-inst, \_& 0.7466 & 0.6172 & \textbackslash{} & \textbackslash{} & 6    & 6   & \textbackslash{} & \textbackslash{} \\

& \_, MI& 0.7935 & 0.6514 & 0.3951 & 0.5885 & 0  & 0  & 7  & 3   \\
\midrule

\multirow{5}{*}{Output}&Natural, TxtLabel, PU & 0.8113 & 0.7060 & 0.4709 & 0.6365 & 0 & 4 & 13 & 18  \\

&Lines, \_, \_ & 0.8103 & 0.7057 & 0.4691 & 0.6387 & 0  & 0  & 3  & 0   \\

&JSON, \_, \_& 0.8118 & 0.7064 & 0.5237 & 0.6323 & 0  & 0  & 1  & 0   \\

& \_, NumLabel, \_ & 0.8121 & 0.6962 & 0.4042 & 0.6697 & 10 & 17 & 4  & 15  \\

& \_, \_, OU& 0.8061 & 0.6467 & 0.4843 & 0.5155 & 1  & 25 & 44 & 4 \\
\midrule

\multirow{3}{*}{Reasoning}&No-CoT & 0.8118 & 0.7064 & 0.5237 & 0.6323 & 0  & 0  & 1  & 0   \\

&CoT& 0.7995 & 0.7026 & 0.4992 & 0.6975 & 2273 & 193 & 560 & 2043 \\

&R-CoT& 0.8087 & 0.6961 & 0.5022 & 0.6772 & 57 & 48  & 85 & 167 \\    

\bottomrule[1.2pt]
\end{tabular}
\caption{MASA evaluations of each SDE option for model \textbf{bc2-chat}. Definition of "\_" see Table \ref{fig:LLaMA2-Chat_result}.}
\label{fig:Baichuan2-Chat_result}
\end{table*}
\begin{table*}[htbp]
\centering
\small
\renewcommand{\arraystretch}{0.9}
\setlength\tabcolsep{4pt}
\begin{tabular}{ll|cc|cc|cc|cc}
\toprule[1.2pt]
   \multicolumn{2}{c|}{model: \textbf{bc2-base}} &\multicolumn{4}{c|}{\textbf{Weighted Kappa $\kappa$}} &\multicolumn{4}{c}{\textbf{\# Wrong format} {\scriptsize(7969 test samples in total)}}  \\
\midrule
  \multicolumn{2}{c|}{train\_size=500} & D1→D1 & D2→D2 & D1→D2 & D2→D1  & D1→D1 & D2→D2 & D1→D2 & D2→D1    \\
\midrule

\multirow{4}{*}{Input}&Inst-last, No-MI & 0.8017 & 0.6412 & 0.4441 & 0.6146 & 0  & 0  & 75  & 0   \\

&Inst-first, \_ & 0.8016 & 0.6649 & 0.4488 & 0.6657 & 0 & 6 & 27 & 4   \\

&No-inst, \_& 0.7533 & 0.6020 & \textbackslash{} & \textbackslash{} & 2  & 3  & \textbackslash{} & \textbackslash{} \\

& \_, MI& 0.7660 & 0.4999 & 0.3220 & 0.1978 & 0  & 0  & 1  & 164 \\
\midrule

\multirow{5}{*}{Output}&Natural, TxtLabel, PU & 0.8017 & 0.6412 & 0.4441 & 0.6146 & 0  & 0  & 75  & 0   \\

&Lines, \_, \_ & 0.7996 & 0.6317 & 0.4583 & 0.6191 & 0  & 0  & 2   & 0   \\

&JSON, \_, \_ & 0.8008 & 0.6476 & 0.4316 & 0.6104 & 0  & 0  & 0   & 0   \\

& \_, NumLabel, \_& 0.7969 & 0.5794 & 0.4312 & 0.5206 & 7  & 45 & 469 & 47  \\

& \_, \_, OU& 0.7595 & 0.5202 & 0.4240 & 0.4944 & 0  & 0  & 116 & 2   \\
\midrule

\multirow{3}{*}{Reasoning}&No-CoT  & 0.7595 & 0.5202 & 0.4240 & 0.4944 & 0  & 0  & 116 & 2   \\
&CoT& 0.7865 & 0.6814 & 0.3854 & 0.6745 & 63 & 17 & 43  & 483 \\

&R-CoT& 0.7980 & 0.6548 & 0.4240 & 0.6349  & 32 & 44 & 39  & 32 \\

\midrule[1.2pt]
  \multicolumn{2}{c|}{train\_size=1000} & D1→D1 & D2→D2 & D1→D2 & D2→D1  & D1→D1 & D2→D2 & D1→D2 & D2→D1    \\
\midrule

\multirow{4}{*}{Input}&Inst-last, No-MI& 0.8143 & 0.6981 & 0.4747 & 0.6767 & 0  & 0  & 26  & 4   \\
&Inst-first, \_ & 0.8155 & 0.7157 & 0.5061 & 0.6974 & 0  & 3  & 26  & 4   \\

&No-inst, \_& 0.7543 & 0.6391 & \textbackslash{} & \textbackslash{} & 0  & 3  & \textbackslash{} & \textbackslash{} \\
& \_, MI& 0.8010 & 0.6489 & 0.4164 & 0.5250 & 0  & 0  & 1   & 431 \\
\midrule

\multirow{5}{*}{Output}&Natural, TxtLabel, PU & 0.8143 & 0.6981 & 0.4747 & 0.6767 & 0  & 0  & 26  & 4  \\

&Lines, \_, \_ & 0.8103 & 0.7003 & 0.4732 & 0.6713 & 0  & 0  & 6 & 1  \\

&JSON, \_, \_ & 0.8120 & 0.7039 & 0.4785 & 0.6819 & 0  & 0  & 0   & 0   \\

& \_, NumLabel, \_ & 0.8119 & 0.6812 & 0.4575 & 0.6467 & 1 & 5 & 292 & 8   \\

& \_, \_, OU& 0.7894 & 0.6484 & 0.4031 & 0.6235 & 0  & 1  & 31  & 0   \\
\midrule

\multirow{3}{*}{Reasoning}&No-CoT & 0.7894 & 0.6484 & 0.4031 & 0.6235 & 0 & 1 & 31 & 0   \\

&CoT& 0.8045 & 0.7063 & 0.5319 & 0.6965 & 21 & 12 & 25  & 494 \\

&R-CoT& 0.8160 & 0.7021 & 0.4604 & 0.6949 & 15 & 14 & 24 & 115\\

\bottomrule[1.2pt]
\end{tabular}
\caption{MASA evaluations of each SDE option for model \textbf{bc2-base}. Definition of "\_" see Table \ref{fig:LLaMA2-Chat_result}.}
\label{fig:Baichuan2-Base_result}
\end{table*}

\subsection{Detailed Results on GENIA, MAVEN and Review11}\label{app:experiment2}
Table \ref{tab:detailed-3downstream-exp} shows the comparison of different sample design strategies on three downstream tasks—GENIA (Nested NER), MAVEN (Event Detection), and Review11 (MASA). 
Hard and soft-matching F1 scores are reported for GENIA and MAVEN, while kappa $\kappa$ and accuracy are reported for Review11.
From the results, we can see that ES-SDE maintains its advantage over other methods, across different tasks and training sizes.

Table \ref{tab:inst-var} illustrates the performances of different sample design strategies on three downstream tasks across different instruction variations.

\begin{table*}[]
\setlength\tabcolsep{8pt}
\centering
\begin{tabular}{c|c|cc|cc|cc}
\toprule
                      &                            & \multicolumn{2}{c|}{\textbf{GENIA} (Nested-NER)} & \multicolumn{2}{c|}{\textbf{MAVEN}   (ED)}  & \multicolumn{2}{c}{\textbf{Review11} (MASA)} \\
training size       & {\small\textbf{Strategies}} & F1-hard            & F1-soft           & F1-hard         & F1-soft         & $\kappa$            & Acc              \\ \midrule
\multirow{3}{*}{$500$}  & heuristic                  & 0.51232            & 0.57465           & 0.5197          & 0.5356          & 0.588            & 0.7586           \\
                      & EW-SDE                     & 0.48328            & 0.54318           & 0.4922          & 0.5364          & 0.7235           & 0.8327           \\
                      & \textbf{ES-SDE}            & \textbf{0.54068}   & \textbf{0.61412}  & \textbf{0.5846} & \textbf{0.6331} & \textbf{0.7691}  & \textbf{0.8626}  \\ \midrule
\multirow{3}{*}{$1,000$} & heuristic                  & 0.56537            & 0.62275           & 0.6237          & 0.6354          & 0.7058           & 0.8262           \\
                      & EW-SDE                     & 0.48785            & 0.55166           & 0.6109          & 0.6275          & 0.7565           & 0.8502           \\
                      & \textbf{ES-SDE}            & \textbf{0.61593}   & \textbf{0.68951}  & \textbf{0.6432} & \textbf{0.6726} & \textbf{0.7892}  & \textbf{0.8716}  \\ \midrule
\multirow{3}{*}{$2,000$} & heuristic                  & 0.64759            & 0.69905           & 0.6722          & 0.6813          & 0.7479           & 0.8483           \\
                      & EW-SDE                     & 0.54351            & 0.6025            & 0.6966          & 0.7106          & 0.7805           & 0.8649           \\
                      & \textbf{ES-SDE}            & \textbf{0.68069}   & \textbf{0.7393}   & \textbf{0.7033} & \textbf{0.7172} & \textbf{0.8023}  & \textbf{0.8785}  \\ \midrule
\multirow{3}{*}{$4,000$} & heuristic                  & 0.68726            & 0.73825           & 0.7118          & 0.7176          & 0.7751           & 0.8644           \\
                      & EW-SDE                     & 0.71109            & 0.77093           & 0.7265          & 0.7338          & 0.7917           & 0.8715           \\
                      & \textbf{ES-SDE}            & \textbf{0.72726}   & \textbf{0.78487}  & \textbf{0.7295} & \textbf{0.7466} & \textbf{0.805}   & \textbf{0.8814}  \\ \bottomrule
\end{tabular}
\caption{Comparison of different sample design strategies on three downstream tasks. ES-SDE maintains its advantage over other methods, across different tasks and training sizes.}
\label{tab:detailed-3downstream-exp}
\end{table*}
\begin{table*}[]
\setlength\tabcolsep{4pt}
\centering
\begin{tabular}{c|c|cc|cc|cc}
\toprule
   &                 & \multicolumn{2}{c|}{\textbf{GENIA} (Nested-NER)} & \multicolumn{2}{c|}{\textbf{MAVEN} (ED)}    & \multicolumn{2}{c}{\textbf{Review11} (MASA)} \\
 Instruction Variation  &    \textbf{Strategies}             & F1-hard            & F1-soft           & F1-hard         & F1-soft         & $\kappa$            & Acc              \\ \midrule
\multirow{3}{*}{\textbf{inst-1}}    & heuristic       & 0.5123             & 0.5747            & 0.5197          & 0.5356          & 0.588            & 0.7586           \\
                                        & EW-SDE          & 0.4833             & 0.5432            & 0.4922          & 0.5364          & 0.7235           & 0.8327           \\
                                        & \textbf{ES-SDE} & \textbf{0.5407}    & \textbf{0.6141}   & \textbf{0.5846} & \textbf{0.6331} & \textbf{0.7691}  & \textbf{0.8626}  \\  \midrule
\multirow{3}{*}{\textbf{inst-2}} & heuristic       & 0.49813            & 0.56095           & 0.5134          & 0.5334          & 0.6009           & 0.7685           \\
           & EW-SDE          & 0.48593            & 0.54999           & 0.4956          & 0.5339          & 0.7208           & 0.8344           \\
          & \textbf{ES-SDE} & \textbf{0.53479}   & \textbf{0.60767}  & \textbf{0.5636} & \textbf{0.6167} & \textbf{0.7659}  & \textbf{0.8615}  \\  \midrule
\multirow{3}{*}{\textbf{inst-3}} & heuristic       & 0.48733            & 0.55491           & 0.4940          & 0.5060          & 0.5793           & 0.7533           \\
                                        & EW-SDE          & 0.47638            & 0.53685           & 0.4925          & 0.5399          & 0.721            & 0.8365           \\
                                        & \textbf{ES-SDE} & \textbf{0.53525}   & \textbf{0.60902}  & \textbf{0.5530} & \textbf{0.6087} & \textbf{0.7624}  & \textbf{0.8601}  \\ \bottomrule
\end{tabular}
\caption{Performances of different sample design strategies on three downstream tasks across different instruction variations.}
\label{tab:inst-var}
\end{table*}

\subsection{Additional Analysis on \textit{Inst-last} and \textit{Inst-first}}\label{app:input}
\begin{figure*}
    \centering
    \includegraphics[width=1\linewidth]{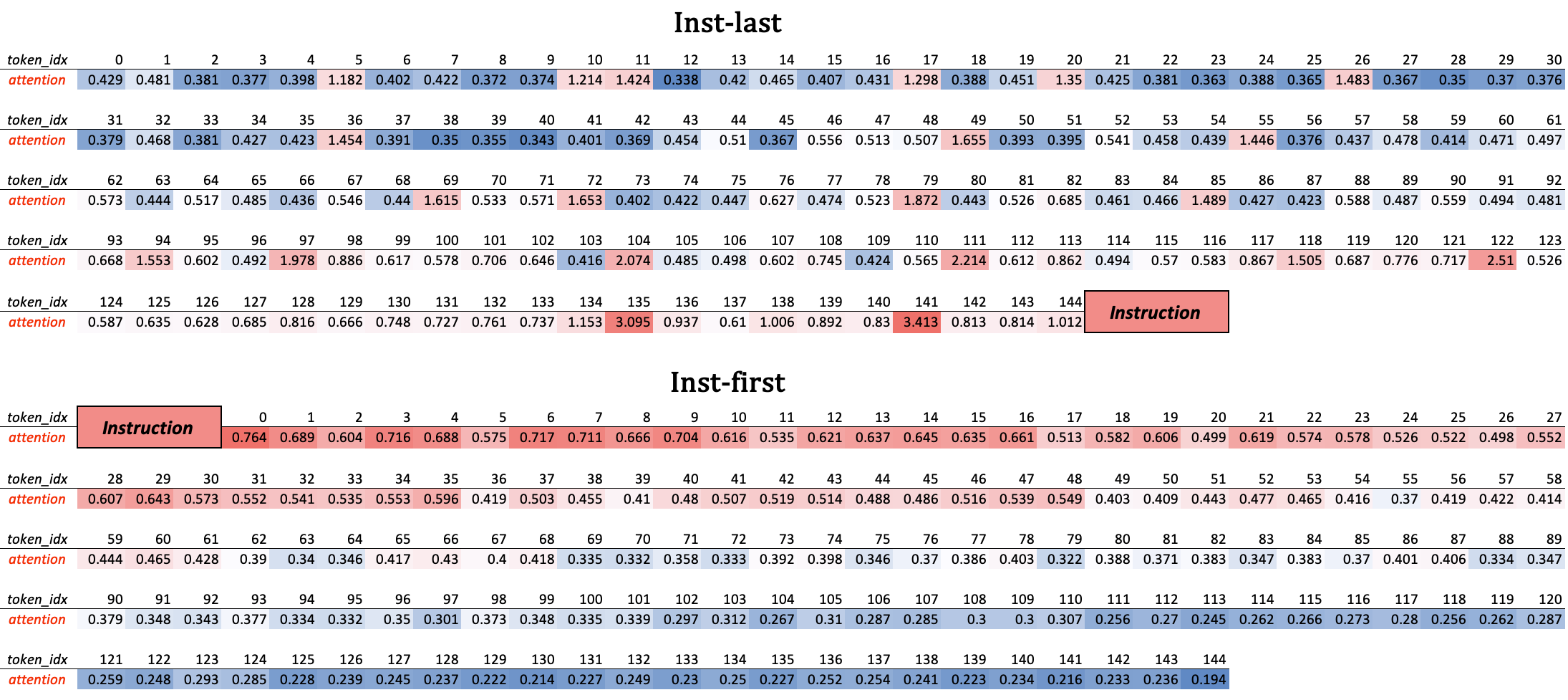}
    \caption{Comparison of task-related attention scores using \textit{Inst-last} and \textit{Inst-first}.}
    \label{fig:inst-attention}
\end{figure*}

The experimental results showing that \textit{Inst-first} consistently outperforms \textit{Inst-last} across various tasks and models are thought-provoking, leading us to conduct a more in-depth analysis. We extract the attention weights related to some task-related fields in the instruction, and sum up these task-related attention weights for each token. Figure \ref{fig:inst-attention} shows the comparison of the attention weights for a certain customer review. As we can see, \textbf{tokens that are closer to the instruction usually get higher task-related attention weights}. Intuitively, when people write reviews, they generally present their core opinions at the beginning. This leads to the possibility that if the instructions are placed at the front, those core parts may receive greater task-related attention weights. This may partly explain why \textit{Inst-first} usually leads to a higher sentiment analysis performance.

\subsection{Additional Analysis on \textit{OU} and \textit{PU}}\label{app:ou_pu}
In previous experiments, we found that \textit{OU} performs much worse than \textit{PU}. This intriguing result motivates us to a further analysis. Specifically, we calculate and compare the kappa scores of \textit{OU} and \textit{PU} for each aspect, to analyze the relationship between label distributions and the effect of \textit{OU}.

From the result in Table \ref{table:mentioned-only}, we can observe that when training the model with 500 samples, for aspects with a higher number of unmentioned, the \textit{OU} method showed a significant gap compared to the \textit{PU} format. When the training set increased to 1000 samples, this gap noticeably narrowed. This suggests that for the \textit{OU} method, aspects with more unmentioned, implying less frequent occurrence in answers, are harder for the model to learn, so requiring more data. From another perspective, it also indicates that even if a certain aspect is not covered in the text, mentioning this aspect in the answers can enhance the model's understanding of it.
\begin{table}[h]
\small
\setlength\tabcolsep{1pt}
\renewcommand{\arraystretch}{1.2}
\resizebox{0.5\textwidth}{!}{
\begin{tabular}{ll|c|cc|c|cc}
\toprule
\multicolumn{2}{c|}{\multirow{2}{*}{Aspect}} & \multicolumn{3}{c|}{Trainsize=500} & \multicolumn{3}{c}{Trainsize=1000}  \\
\cline{3-8}
\multicolumn{2}{c|}{}    & (\%)Num\_      & \multicolumn{2}{c|}{$\Delta$$\kappa$} & (\%)Num\_    & \multicolumn{2}{c}{$\Delta$$\kappa$} \\
&    & Unmen   & Avg\_Chat & Avg\_Base & Unmen   & Avg\_Chat & Avg\_Base \\
\hline
D1 & F                       & 1.00   & -.0004      & .0007      & 1.40   & -.0026      & -.0011     \\
   & SA    & 33.60  & -.0687  & -.0555 & 34.40  & -.0062  & -.0212  \\
   & P    & 38.00  & -.0469      & -.0495     & 37.90  & -.0068   & -.0255     \\
   & B    & 65.40  & -.0410       & -.0291     & 65.70  & -.0117   & -.0079  \\
   & H    & 78.00  & -.0920       & -.1367     & 76.40  & -.0033 & -.0207     \\
   & PC    & 93.60  & -.2338      & -.2590     & 93.00  & -.0181      & -.0305     \\
\hline
D2 & TC   & 26.80  & -.0891      & -.1341     & 25.60  & -.0497      & -.0492     \\
   & D                       & 41.60  & -.1106  & -.2475     & 39.10  & -.0280  & -.0500 \\
   & Q    & 61.80  & -.0329 & -.0588     & 61.00  & -.0361      & -.0149     \\
   & SS   & 71.40  & -.2537      & -.2575     & 71.60  & -.0574 & -.0896     \\
   & N     & 94.80  & -.3347      & -.3954     & 93.90  & -.0494      & -.1405  \\
\bottomrule
\end{tabular} }
\caption{Number of ‘Unmentioned’ labels and average $\Delta$$\kappa$ ($\kappa_{OU}$-$\kappa_{PU}$) for different aspects.}
\label{table:mentioned-only}
\end{table}

\subsection{Can PE Guide SDE? Detailed Results}\label{app:pe_analysis}
Evaluating the performances of sample designs involves fine-tuning models on downstream tasks, which can be time-consuming. 
Therefore, we also pondered whether it might be possible to design better samples without training models first. We tried to understand the inherent capabilities and potential of the model by experimenting with different prompt designs in both the zero-shot and in-context learning scenarios.\\
\subsubsection{ Zero-shot and In-context Learning Analysis}
Zero-shot and In-context learning ability can directly reveal LLMs' familiarity with the given task. In the zero-shot approach, we use the input (which contains the instruction on output format) from each SDE option as the prompt for the original frozen LLMs prediction. For the ICL approach, we add two fixed examples from the training set before each test instance. Considering the inference time cost caused by the increase in sample length, we limit our prediction and analysis to 500 samples. All other experimental setups remain aligned with those described in Experiments \uppercase\expandafter{\romannumeral1}. 

\textbf{\textit{Zero-shot Study.}} All six 7B LLMs used in Section \ref{sec:Exp-1} exhibit poor zero-shot MASA ability, failing to follow the instructions to generate proper output in most cases, as shown in Table \ref{table:zero-shot}, making it hard to analysis its relationship with SDE results. Variations in format preferences across different models are observed, which we conjecture is strongly related to the datasets employed for instruction fine-tuning in each model. Some patterns are also contradictory between zero-shot and SDE. For example, the \textit{OU} SDE option consistently harms DT performances, however, its prompts result in notably fewer format errors in zero-shot inference, for certain LLMs. Therefore, zero-shot performances can hardly tell good or bad SDE options.

\textbf{\textit{In-context Learning Study.}} ICL can effectively improve LLMs' instruction-following abilities resulting in far fewer formatting errors than zero-shot. Therefore we report the average sentiment analysis performances of each model on two domains in Table \ref{table:few-shot}. The results suggest that \textit{Inst-first} and \textit{CoT} enhance the performance of most models, which provides valuable insights for format selection during the fine-tuning process. For output designs, \textit{JSON} and \textit{OU} options outperform the other approaches for some models, differing from the SDE results.

\subsubsection{Perplexity Analysis}
Perplexity measures the uncertainty of the model in generating a given text sequence \cite{chen1998evaluation}, with lower perplexity values indicating more confident predictions by the model. In calculations, we estimate perplexity using the common practice of taking the logarithm of the model's loss.

In our task, we compare the PPL scores of the ICL prompts corresponding to each different SDE option, as well as the conditional PPL of the models' ICL predictions. 
For predictions, we concatenate the prompt and the prediction together as a sequence, then consider the prompt as its context.

The perplexity results for different designs are shown in Table \ref{fig:perplexity}.
For input designs, the PPL score of \textit{Inst-first} option is lower than that of \textit{Inst-last} in general, which is consistent with the conclusion that \textit{Inst-first} performs better in ICL and SDE experiments. 
For output designs, the \textit{OU} option gets the highest score, which is inconsistent with its performance on the ICL, but is consistent with its being the worst option in the SDE experiment. Surprisingly, the \textit{JSON} format achieved the significantly lowest ppl score, but it was on par with the \textit{Lines} format in ICL and even worse than \textit{Lines} in SDE.
The most interesting result appears in the reasoning designs. The \textit{CoT} and \textit{R-CoT} options have low PPL scores on prompts but have high scores on predictions conversely. Such contradictions make it difficult to analyze the results of ICL or SDE through PPL scores.

The analysis above also highlights the indispensability of our SDE experiments, cause we cannot predetermine the final effectiveness of different designs through preliminary analysis alone.

\begin{figure*}[t]
    \centering   \includegraphics[width=\textwidth]{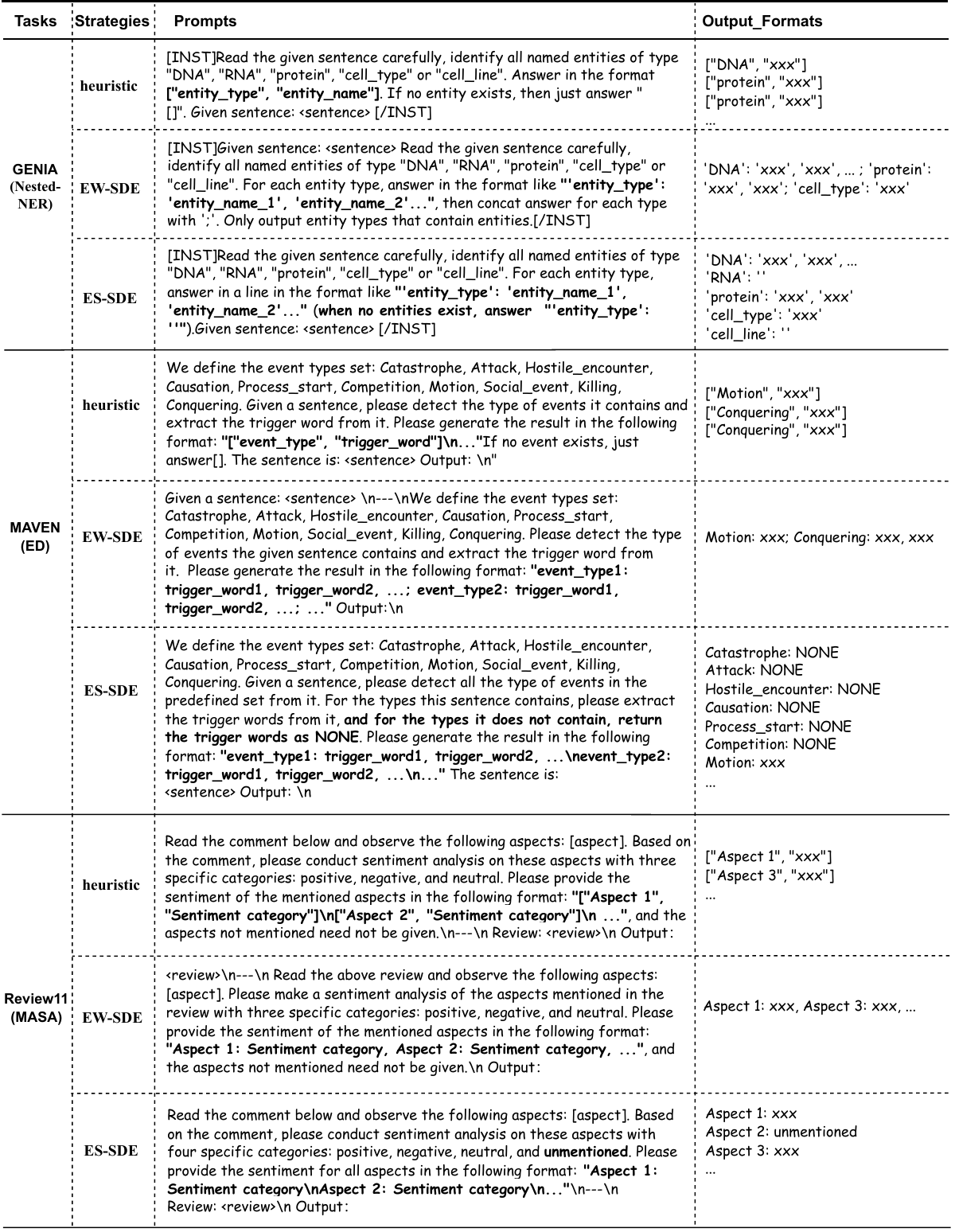}
    \caption{Examples of different sample designs on GENIA, MAVEN and Review11.}
    \label{fig:examples-of-strategies-3new}
\end{figure*}

\begin{figure*}[t]
    \centering   \includegraphics[width=\textwidth]{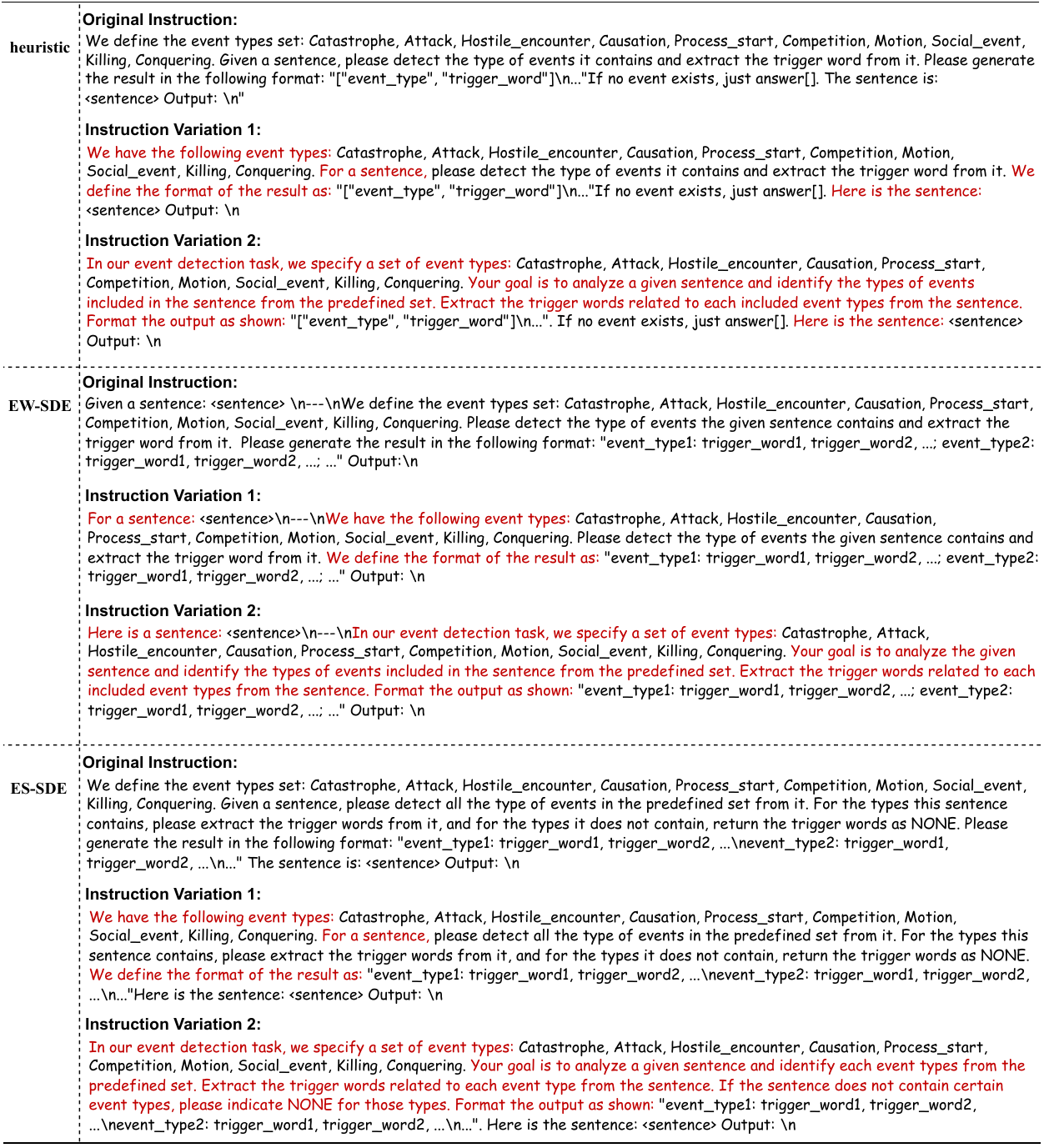}
    \caption{Variations of Instructions on different strategies.(taking MAVEN as an example)}   \label{fig:new_instructions}
\end{figure*}

\begin{table*}[]
\setlength\tabcolsep{5pt}
\centering
\small
\renewcommand{\arraystretch}{0.9}
\begin{tabular}{ll|cccccc}
\toprule[1.2pt]
\multicolumn{2}{c|}{\textbf{Perplexity:Prompts}} & \multicolumn{1}{c}{c-llama2-chat} & \multicolumn{1}{c}{c-llama2-base} & \multicolumn{1}{c}{intern-chat} & \multicolumn{1}{c}{intern-base} & \multicolumn{1}{c}{bc2-chat} & \multicolumn{1}{c}{bc2-base} \\
\midrule
\multirow{2}{*}{Input} & Inst-last, No-MI & 47.662 & 111.063 & 18.422 & 19.036 & 59.046 & 42.030 \\
 & Inst-first, \_ & 46.357 & 110.065 & 19.561 & 18.632 & 54.795 & 39.003 \\
\midrule
\multirow{5}{*}{Output} & Natural, TxtLabel, PU & 47.662 & 111.063 & 18.422 & 19.036 & 59.046 & 42.030 \\
 & Lines, \_, \_ & 47.918 & 191.274 & 18.561 & 19.219 & 60.498 & 42.638 \\
 & JSON, \_, \_ & 29.008 & 78.848 & 14.675 & 13.260 & 38.547 & 25.405 \\
 & \_, NumLabel, \_ & 41.690 & 92.717 & 17.664 & 16.348 & 51.963 & 35.185 \\
 & \_, \_, OU & 55.345 & 129.055 & 20.862 & 21.450 & 69.022 & 49.426 \\
\midrule
\multirow{3}{*}{Reasoning} & No-CoT & 29.008 & 78.848 & 14.675 & 13.260 & 38.547 & 25.405 \\
 & CoT & 18.263 & 41.312 & 10.812 & 9.379 & 23.406 & 15.267 \\
 & R-CoT & 18.210 & 42.648 & 10.789 & 9.354 & 22.671 & 15.333\\

\midrule[1.2pt]
\multicolumn{2}{c|}{\textbf{Perplexity:Predictions}} & \multicolumn{1}{c}{c-llama2-chat} & \multicolumn{1}{c}{c-llama2-base} & \multicolumn{1}{c}{intern-chat} & \multicolumn{1}{c}{intern-base} & \multicolumn{1}{c}{bc2-chat} & \multicolumn{1}{c}{bc2-base} \\
\midrule
\multirow{2}{*}{Input} & Inst-last, No-MI & 1.052 & 1.109 & 1.051 & 1.394 & 1.061 & 1.127 \\
 & Inst-first, \_ & 1.088 & 1.284 & 1.046 & 1.360 & 1.066 & 1.113 \\
\midrule
\multirow{5}{*}{Output} & Natural, TxtLabel, PU & 1.052 & 1.109 & 1.051 & 1.394 & 1.061 & 1.127 \\
 & Lines, \_, \_ & 1.052 & 1.137 & 1.058 & 1.386 & 1.222 & 1.136 \\
 & JSON, \_, \_ & 1.038 & 1.074 & 1.045 & 1.407 & 1.019 & 1.042 \\
 & \_, NumLabel, \_ & 1.096 & 1.142 & 1.078 & 1.403 & 1.088 & 1.102 \\
 & \_, \_, OU & 1.183 & 1.368 & 1.089 & 1.279 & 1.353 & 1.823 \\
\midrule
\multirow{3}{*}{Reasoning} & No-CoT & 1.038 & 1.074 & 1.045 & 1.407 & 1.019 & 1.042 \\
 & CoT & 1.234 & 1.475 & 1.084 & 1.186 & 1.090 & 1.129 \\
 & R-CoT & 1.239 & 1.293 & 1.069 & 1.185 & 1.063 & 1.090 \\

\bottomrule[1.2pt]
\end{tabular}
\caption{The PPL scores on the ICL prompts and predictions corresponding to each SDE options on the MASA ID tasks.}
\label{fig:perplexity}
\end{table*}
\begin{table*}[!ht]
\resizebox{\textwidth}{!}{%
\begin{tabular}{ll|cc|cc|cc|cc|cc|cc}
\toprule[1.2pt]
 & &
  \multicolumn{2}{c|}{\textbf{c-llama2-chat}} &
  \multicolumn{2}{c|}{\textbf{Intern-chat}} &
  \multicolumn{2}{c|}{\textbf{bc2-chat}} &
  \multicolumn{2}{c|}{\textbf{c-llama2-base}} &
  \multicolumn{2}{c|}{\textbf{Intern-base}} &
  \multicolumn{2}{c}{\textbf{bc2-base}} \\
 & &
  \multicolumn{1}{c}{D1} &
  \multicolumn{1}{c|}{D2} &
  \multicolumn{1}{c}{D1} &
  \multicolumn{1}{c|}{D2} &
  \multicolumn{1}{c}{D1} &
  \multicolumn{1}{c|}{D2} &
  \multicolumn{1}{c}{D1} &
  \multicolumn{1}{c|}{D2} &
  \multicolumn{1}{c}{D1} &
  \multicolumn{1}{c|}{D2} &
  \multicolumn{1}{c}{D1} &
  \multicolumn{1}{c}{D2} \\ 
\midrule
\multirow{2}{*}{Input} &
Ins-last &
  74.24 &
  31.67 &
  85.82 &
  11.75 &
  40.67 &
  22.12 &
  88.92 &
  36.60 &
  94.89 &
  81.60 &
  100 &
  98.18 \\
& Ins-first &
  70.05 &
  44.82 &
  98.76 &
  99.61 &
  59.56 &
  24.18 &
  88.62 &
  27.49 &
  89.79 &
  75.59 &
  99.66 &
  96.26 \\ 
\midrule
\multirow{5}{*}{Output} &
Natural, TxtLabel, PU &
  74.24 &
  31.67 &
  85.82 &
  11.75 &
  40.67 &
  22.12 &
  88.92 &
  36.60 &
  94.89 &
  81.60 &
  100 &
  98.18 \\
& Lines, \_, \_ &
  1.18 &
  1.31 &
  99.94 &
  97.06 &
  4.17 &
  1.57 &
  72.51 &
  12.10 &
  99.57 &
  99.79 &
  99.99 &
  99.94 \\
& JSON, \_, \_ &
  5.94 &
  16.49 &
  100 &
  100 &
  96.15 &
  73.53 &
  99.94 &
  100 &
  100 &
  100 &
  100 &
  100 \\ 
& \_, Numerical, \_ &
  99.87 &
  92.21 &
  99.99 &
  100 &
  100 &
  100 &
  100 &
  100 &
  100 &
  100 &
  100 &
  100 \\
& \_, \_, OU &
  45.75 &
  18.31 &
  70.21 &
  31.38 &
  44.15 &
  50.93 &
  72.79 &
  87.99 &
  76.80 &
  56.87 &
  99.74 &
  95.33 \\ 
\midrule
\multirow{2}{*}{Reasoning} &
No-CoT &
  5.94 &
  16.49 &
  100 &
  100 &
  96.15 &
  73.53 &
  99.94 &
  100 &
  100 &
  100 &
  100 &
  100 \\
& CoT &
  35.25 &
  34.25 &
  100 &
  100 &
  58.66 &
  53.29 &
  100 &
  100 &
  100 &
  100 &
  99.99 &
  99.99 \\
& R-CoT &
  33.84 &
  75.87 &
  100 &
  100 &
  80.71 &
  77.12 &
  98.24 &
  90.58 &
  100 &
  100 &
  100 &
  100 \\ 
\bottomrule[1.2pt]
\end{tabular}%
}
\caption{Format error rate(\%) in zero-shot scenario}
\label{table:zero-shot}
\end{table*}
\begin{table*}[p]
\centering
\small
\renewcommand{\arraystretch}{0.9}
\setlength\tabcolsep{4pt}
\begin{tabular}{ll|cc|cc|cc}
\toprule[1.2pt]

\multicolumn{2}{c|}{test\_size=500} & \textbf{c-llama2-chat} & \textbf{c-llama2-base} & \textbf{intern-chat} & \textbf{intern-base} & \textbf{bc2-chat} & \textbf{bc2-base} \\
\midrule
 
\multirow{2}{*}{Input}  & Inst-last  & 0.3834 & 0.2835 & 0.1856 & 0.1212 & 0.4402 & 0.4187 \\
& Inst-first & 0.4832 & 0.2959 & 0.2038 & 0.2044  & 0.5091 & 0.4345 \\
\midrule

\multirow{5}{*}{Output} & Natural, TxtLabel, PU   & 0.3834 & 0.2835 & 0.1856 & 0.1212 & 0.4402 & 0.4187 \\
& Lines, \_, \_     & 0.4220 & 0.2921 & 0.2436 & 0.1846  & 0.3971 & 0.4077 \\
& JSON, \_, \_      & 0.3773 & 0.2132 & 0.3390 & 0.2954  & 0.4614 & 0.3683 \\
& \_, NumLabel, \_ & 0.1522 & 0.1666 & 0.2470 & 0.2603  & 0.2406 & 0.1960 \\
& \_, \_, OU        & 0.3612 & 0.3168 & 0.2461 & 0.1443  & 0.1948 & 0.1924 \\
\midrule

\multirow{3}{*}{Reasoning}    & No-CoT    & 0.3773 & 0.2132 & 0.3390 & 0.2954  & 0.4614 & 0.3683 \\
& CoT       & 0.3383 & 0.2174 & 0.3636 & 0.3167  & 0.4810 & 0.4466 \\
& R-CoT   & 0.3638 & 0.2445 & 0.3522 & 0.2633  & 0.4668 & 0.4075\\

\bottomrule[1.2pt]
\end{tabular}
\caption{The average weighted Kappa $\kappa$ on the MASA ID tasks in in-context learning scenario}
\label{table:few-shot}
\end{table*}

\end{document}